\pdfoutput=1

\documentclass[10pt,journal,compsoc]{IEEEtran}
\newcommand{\eg}{e.g.\@\xspace}
\newcommand{\ie}{i.e.\@\xspace}
\newcommand{\etal}{\textit{et al}.}

\usepackage{graphicx}
\usepackage{subfig}
\usepackage{color}
\usepackage{array}
\usepackage{booktabs}
\usepackage{textcomp}
\usepackage{url}
\usepackage{wrapfig}
\usepackage{xspace}

\ifCLASSOPTIONcompsoc
  \usepackage[nocompress]{cite}
\else
  \usepackage{cite}
\fi
\ifCLASSINFOpdf
\else
\fi
\begin{document}
%
\title{Incremental Learning Through Deep Adaptation}
%
%
%
%

\author{Amir Rosenfeld\qquad John K. Tsotsos\\
Department of Electrical Engineering and Computer Science\\ York University, Toronto, ON, Canada\\
\texttt{amir@eecs.yorku.ca,tsotsos@cse.yorku.ca}}

%
%

\markboth{}%
{Shell \MakeLowercase{\textit{et al.}}: Bare Demo of IEEEtran.cls for Computer Society Journals}
%



\IEEEtitleabstractindextext{%
\begin{abstract}
Given an existing trained neural network, it is often desirable to
learn new capabilities without hindering performance of those already
learned. Existing approaches either learn sub-optimal solutions, require
joint training, or incur a substantial increment in the number of
parameters for each added domain, typically as many as the original
network. We propose a method called \emph{Deep Adaptation Networks}
(DAN) that constrains newly learned filters to be linear combinations
of existing ones. DANs precisely preserve performance on the original
domain, require a fraction (typically 13\%, dependent on network architecture)
of the number of parameters compared to standard fine-tuning procedures
and converge in less cycles of training to a comparable or better
level of performance. When coupled with standard network quantization
techniques, we further reduce the parameter cost to around 3\% of
the original with negligible or no loss in accuracy. The learned architecture
can be controlled to switch between various learned representations,
enabling a single network to solve a task from multiple different
domains. We conduct extensive experiments showing the effectiveness
of our method on a range of image classification tasks and explore
different aspects of its behavior.
\end{abstract}

}

\maketitle

\IEEEdisplaynontitleabstractindextext

%
\IEEEpeerreviewmaketitle

\IEEEraisesectionheading{\section{Introduction}\label{sec:introduction}}

\begin{figure*}
\includegraphics[width=1\textwidth]{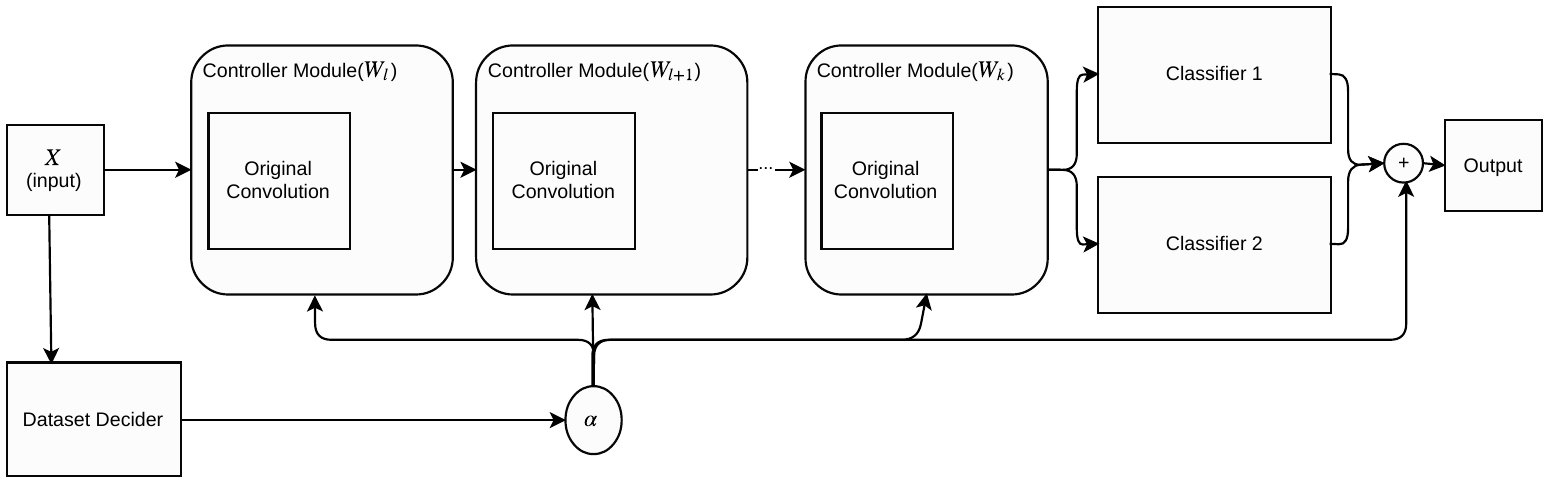}

\caption{\label{fig:Summary-of-proposed} Overview of proposed method. For
newly learned domains, controller modules are attached to convolutions
of a base network, whose parameter are frozen. A switching variable
$\alpha$ allows to switch the behavior of the network between the
original behaviour of the convolution and a re-parametrized one for
the new domain. $\alpha$ can be determined either manually or via
a sub-network (``Dataset Decider'') which determines the source
domain of the image, switching accordingly between different sets
of control parameters. $\alpha$ Also controls which of the classifiers
to apply. Other layers (e.g, non-linearities, batch-normalization,
skip layers) not shown for presentation purposes. We visualize one
added task, though an arbitrary number of tasks can be added.}
\end{figure*}

\begin{figure}
\begin{centering}
\includegraphics[height=0.35\textwidth]{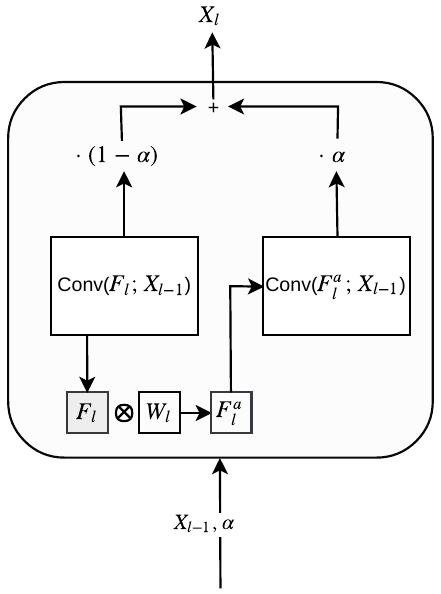}
\par\end{centering}
\caption{\label{fig:Controller-modules}Controller modules: filters of a convolutional
layer of a base network are modified by re-combining their weights
through a controller module, where a switching $\alpha$ variable
can choose between the original filters $F_{l}$ and newly created
ones $F_{l}^{a}$. We show a controller module for a single added
task, but any number of controllers can be added, with $\alpha$ then
being a vector instead of scalar. }

\end{figure}

While deep neural networks continue to show remarkable performance
gains in various areas such as image classification \cite{krizhevsky2012imagenet},
semantic segmentation \cite{long2015fully}, object detection \cite{girshick2014rich},
speech recognition \cite{hannun2014deep} medical image analysis \cite{litjens2017survey}
- and many more - it is still the case that typically, a separate
model needs to be trained for each new task. Given two tasks of a
totally different modality or nature, such as predicting the next
word in a sequence of words versus predicting the class of an object
in an image, it stands to reason that each would require a different
architecture or computation. A more restricted scenario - and the
one we aim to tackle in this work - is that of learning a representation
which works well on several related domains. Such a scenario was recently
coined by Rebuffi \etal \cite{rebuffi2017learning} as multiple-domain
learning (MDL) to set it aside from multi-task learning - where different
tasks are to be performed on the same domain. An example of MDL is
image-classification where the images may belong to different domains,
such as drawings, natural images, etc. In such a setting it is natural
to expect that solutions will:
\begin{enumerate}
\item Utilize the same computational pipeline
\item Require a modest increment in the number of required parameters for
each added domain
\item Retain performance of already learned datasets (avoid ``catastrophic
forgetting'')
\item Be learned incrementally, dropping the requirement for joint training
such as in cases where the training data for previously learned tasks
is no longer available.
\end{enumerate}
Our goal is to enable a network to learn a set of related tasks one
by one while adhering to the above requirements. We do so by augmenting
a network learned for one task with \emph{controller modules} which
utilize already learned representations for another. The parameters
of the controller modules are optimized to minimize a loss on a new
task. The training data for the original task is not required at this
stage. The network's output on the original task data stays exactly
as it was; any number of controller modules may be added to each layer
so that a single network can simultaneously encode multiple distinct
domains, where the transition from one domain to another can be done
by setting a binary switching variable or controlled automatically.
The resultant architecture is coined DAN, standing for \textbf{D}eep
\textbf{A}daptation \textbf{N}etworks. We demonstrate the effectiveness
of our method on the recently introduced Visual Decathlon Challenge
\cite{rebuffi2017learning} whose task is to produce a classifier
to work well on ten different image classification datasets. Though
adding only 13\% of the number of original parameters for each newly
learned task (the specific number depends on the network architecture),
the average performance surpasses that of fine tuning \emph{all }parameters
- without the negative side effects of doubling the number of parameters
and catastrophic forgetting. In this work, we focus on image classification
in various datasets, hence in our experiments the word ``task''
refers to a specific dataset/domain. The proposed method is extensible
to multi-task learning, such as a single network that performs both
image segmentation and classification, but this paper does not pursue
this.

Our main contribution is the introduction of an improved alternative
to transfer learning, which is as effective as fine-tuning all network
parameters towards a new task, precisely preserves old task performance,
requires a fraction (network dependent, typically 13\%) of the cost
in terms of new weights and is able to switch between any number of
learned tasks. Experimental results verify the applicability of the
method to a wide range of image-classification datasets. 

We introduce two variants of the method, a fully-parametrized version,
whose merits are described above and one with far fewer parameters,
which significantly outperforms shallow transfer learning (\ie feature
extraction) for a comparable number of parameters. In the next section,
we review some related work. Sec. \ref{sec:Approach} details the
proposed method. In Sec. \ref{sec:Experiments} we present various
experiments, including comparison to related methods, as well as exploring
various strategies on how to make our method more effective, followed
by some discussion \& concluding remarks.

\section{Related Work}

\subsection{Multi-task Learning}

In multi-task learning, the goal is to train one network to perform
several tasks simultaneously on the same input. This is usually done
by jointly training on all tasks. Such training is advantageous in
that a single representation is used for all tasks. In addition, multiple
losses are said to act as an additional regularizer. Some examples
include facial landmark localization \cite{conf/cvpr/ZhuLLT15}, semantic
segmentation \cite{he2017mask}, 3D-reasoning \cite{eigen2015predicting},
object and part detection \cite{bilen16integrated} and others. While
all of these learn to perform different tasks on the same dataset,
the recent work of \cite{1701.07275} explores the ability of a single
network to perform tasks on various image classification datasets.
We also aim to classify images from multiple datasets but we propose
doing so in a manner which learns them one-by-one rather than jointly.
Concurrent with our method is that of \cite{rebuffi2017learning}
which introduces dataset-specific additional residual units. We compare
to this work in Sec \ref{sec:Experiments}. Our work bears some resemblance
to \cite{journals/corr/MisraSGH16}, where two networks are trained
jointly, with additional ``cross-stitch'' units, allowing each layer
from one network to have as additional input linear combinations of
outputs from a lower layer in another. However, our method does not
require joint training and requires significantly fewer parameters. 

\subsection{Incremental Learning}

Adding a new ability to a neural net often results in so-called ``catastrophic
forgetting'' \cite{french1999catastrophic}, hindering the network's
ability to perform well on old tasks. The simplest way to overcome
this is by fixing all parameters of the network and using its penultimate
layer as a feature extractor, upon which a classifier may be trained
\cite{donahue2014decaf,sharif2014cnn}. While guaranteed to leave
the old performance unaltered, it is observed to yield results which
are substantially inferior to fine-tuning the entire architecture
\cite{girshick2014rich}. The work of Li and Hoiem \cite{journals/corr/LiH16e}
provides a succinct taxonomy of several variants of such methods.
In addition, they propose a mechanism of fine-tuning the entire network
while making sure to preserve old-task performance by incorporating
a loss function which encourages the output of the old features to
remain constant on newly introduced data. While their method adds
a very small number of parameters for each new task, it does not guarantee
that the model retains its full ability on the old task. Rusu \etal
\cite{rusu2016progressive} shows new representations can be added
alongside old ones while leaving the old task performance unaffected.
However, this comes at a cost of duplicating the number of parameters
of the original network for each added task. In Kirkpatrick \etal
\cite{kirkpatrick2017overcoming} the learning rate of neurons is
lowered if they are found to be important to the old task. Our method
\emph{fully} preserves the old representation while causing a modest
increase in the number of parameters for each added task. Recently,
Sarwar \etal \cite{1712.02719} have proposed to train a network
incrementally by sharing a subset of early layers and splitting later
ones. By construction, their method also preserves the old representation.
However, as their experiments show, re-learning the parameters of
the new network branches, whether initialized randomly with a similar
distribution or simply copying the old ones results in worse performance
than learning without any network sharing. Our method, while also
re-utilizing existing weights, attains results which are on average
better than simple transfer learning or learning from scratch. 

\subsection{Network Compression}

Multiple works have been published on reducing the number weights
of a neural network as a means to represent it compactly \cite{han2015deep,han2015learning},
gain speedups \cite{denton2014exploiting} or avoid over-fitting \cite{hanson1988comparing},
using combinations of coding, quantization, pruning and tensor decomposition.
Such methods can be used in conjunction with ours to further improve
results, as we show in Sec. \ref{sec:Experiments}.

\begin{figure*}
\begin{centering}
\subfloat[]{\includegraphics[width=1\textwidth]{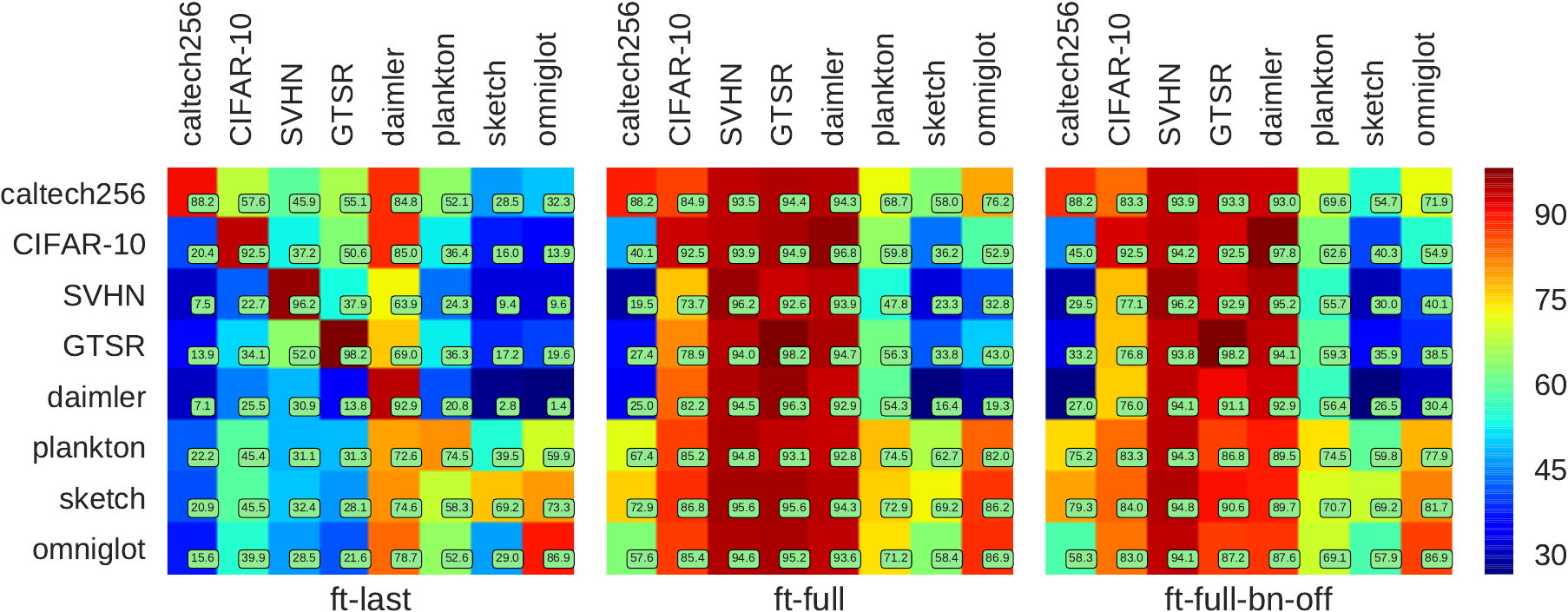}}
\par\end{centering}
\caption{\label{fig:Transferability-of-various} Transferability of various
datasets to each other (\emph{ft-last}) fine tuning only the last
layer (\emph{full}) fine-tuning all layers (\emph{ft-full-bn-off})
fine tuning all layers while disallowing batch-normalization layers'
weights to be updated. Overall, networks tend to be more easily transferable
to problems from related domains (\eg, natural / drawing). Zoom in
to see numbers. It is recommended to view this figure in color on-line. }
\end{figure*}

\section{Approach\label{sec:Approach}}

We begin with some notation. Let $T$ be some task to be learned.
Specifically, we use a deep convolutional neural net (DCNN) in order
to learn a classifier to solve $T$, which is an image classification
task. Most contemporary DCNN's follow a common structure: for each
input $x$, the DCNN computes a representation of the input by passing
it through a set of $l$ layers $\phi_{i}$, $i\in1\ldots l$ interleaved
with non-linearities. The initial (lower) layers of the network are
computational blocks, \eg convolutions with optional residual units
in more recent architectures \cite{he2016deep}. Our method applies
equally to networks with or without residual connections. At least
one fully connected layer $f_{i}$, $i\in1\ldots c$ is attached to
the output of the last convolutional layer. Let $\Phi_{F_{N}}=\sigma(\phi_{l})\circ\ldots\sigma(\phi_{2})\circ\sigma(\phi_{1})$
be the composition of all of the convolutional layers of the network
$N$, interleaved by non-linearities. We use an architecture where
all non-linearities $\sigma$ are the same function, with no tunable
parameters. Denote by $\Phi_{F_{N}}(x)$ the \emph{feature} part of
$N$. Similarly, denote by $\Phi_{C_{N}}=f_{c}\circ\ldots\sigma(f_{2})\circ\sigma(f_{1})$
the \emph{classifier} part of $N$, \ie the composition of all of
the fully-connected layers of $N$. The output of $N$ is then simply
defined as: 
\begin{equation}
N(x)=\Phi_{C_{N}}\circ\Phi_{F_{N}}(x)
\end{equation}

We do not specify batch-normalization layers in the above notation
for brevity. It is possible to drop the $\Phi_{C_{N}}$ term, if the
network is fully convolutional, as in \cite{long2015fully}.

\subsection{Adapting Representations\label{subsec:Adapting-Representations}}

Assume that we are given two tasks, $T_{1}$ and $T_{2}$, to be learned,
and that we have learned a \emph{base network }$N$ to solve $T_{1}$.
We assume that a good solution to $T_{2}$ can be obtained by a network
with the same architecture as $N$ but with different parameters.
We augment $N$ so that it will be able to solve $T_{2}$ as well
by attaching a controller module to each of its convolutional layers.
Each controller module uses the existing weights of the corresponding
layer of $N$ to create new convolutional filters adapted to the new
task $T_{2}$: for each convolutional layer $\phi_{l}$ in $N$, let
$F_{l}\in\mathcal{R}^{C_{o}\times C_{i}\times k\times k}$ be the
set of filters for that layer, where $C_{o}$ is the number of output
features, $C_{l}$ the number of inputs, and $k\times k$ the kernel
size (assuming a square kernel). Denote by $b_{l}\in\mathcal{R}^{C}$
the bias. Denote by $\tilde{F_{l}}\in\mathcal{R}^{C_{o}\times D}$
the matrix whose rows are flattened versions of the filters of $F_{l}$,
where $D=C_{i}\cdot k\cdot k$; let $f\in\mathcal{R}^{C_{i}\times k\times k}$
be a filter from $F_{l}$ whose values are 
\begin{equation}
f^{1}=\left(\begin{array}{ccc}
f_{11}^{1} & \cdots & f_{1k}^{1}\\
 & \ddots\\
 &  & f_{kk}^{1}
\end{array}\right),\cdots,f^{i}=\left(\begin{array}{ccc}
f_{11}^{i} & \cdots & f_{1k}^{i}\\
 & \ddots\\
 &  & f_{kk}^{i}
\end{array}\right)\label{eq:matrix-form}
\end{equation}

The flattened version of $f$ is a row vector:

\begin{equation}
\tilde{f}=(f_{11}^{1},\cdots,f_{kk}^{1},\cdots,\cdots f_{11}^{i},\cdots,f_{kk}^{i})\in\mbox{\ensuremath{\mathcal{R^{D}}}}\label{eq: flattened}
\end{equation}
``Unflattening'' a row vector $\tilde{f}$ reverts it to its tensor
form $f\in\mathcal{R}^{C_{i}\times k\times k}.$ This way, we can
write 

\begin{equation}
\tilde{F_{l}^{a}}=W_{l}\cdot\tilde{F_{l}}
\end{equation}

where $W_{l}\in\mathcal{R}^{C_{o}\times C_{o}}$ is a weight matrix
defining linear combinations of the flattened filters of $F_{l}$,
resulting in $C_{o}$ new filters. Unflattening $\tilde{F_{l}^{a}}$
to its original shape results in $F_{l}^{a}\in\mathcal{R}^{C_{o}\times C_{i}\times k\times k}$,
which we call the adapted filters of layer $\phi_{l}$. Using the
symbol $X\otimes Y$ as shorthand for \emph{flatten} $Y$$\rightarrow$\emph{matrix}
\emph{multiply} \emph{by} $X$$\rightarrow$\emph{unflatten}, we can
write: 
\begin{equation}
F_{l}^{a}=W_{l}\otimes F_{l}
\end{equation}

If the convolution contains a bias, we instantiate a new weight vector
$b_{l}^{a}$ instead of the original $b_{l}$. The output of layer
$\phi_{l}$ is computed as follows: let $x_{l}$ be the input of $\phi_{l}$
in the adapted network. For a given switching parameter $\alpha\in\{0,1\}$,
we set the output of the modified layer to be the application of the
switched convolution parameters and biases: 

\begin{equation}
x_{l+1}=[\alpha(W_{l}\otimes F_{l})+(1-\alpha)F_{l}]\ast x_{l}+\alpha b_{l}^{a}+(1-\alpha)b_{l}\label{eq: interpolate}
\end{equation}

The above formulation is for switching between two different behaviours
- that of a base network and that of the controller modules learned
for a new task. 

To allow the network to perform multiple tasks, we turn $\alpha$
to a vector $\alpha\in\{0,1\}^{n}$ where $n$ is the overall number
of tasks, so that $\alpha_{j}=1$if we want to perform the $j$'th
task and 0 otherwise. The output of the $l$ layer will be then determined
similarly to Equation \ref{eq: interpolate}:
\[
x_{l+1}={\ensuremath{\sum}}_{i=1}^{n}\alpha_{i}(F_{l}^{a_{i}}\ast x_{l}+b_{l}^{i})
\]
 Where $F_{l}^{a_{i}}$, $b_{l}^{i}$ are the set of adapted filters
/ bias for the $i$'th task and we define for $F_{l}^{a_{1}}=F_{l},b_{l}^{1}=b_{l}$
to include the original task. 

A set of fully connected layers $f_{i}^{a}$ are learned from scratch,
attaching a new ``head'' to the network for each new task. Throughout
training \& testing, the weights of $F$ (the filters of $N$) are
kept fixed and serve as basis functions for $F^{a}$. The weights
of the controller modules are learned via back-propagation given the
loss function. Weights of any batch normalization (BN) layers are
either kept fixed or learned anew. The batch-normalized output is
switched between the values of the old and new BN layers, similarly
to Eq. \ref{eq: interpolate}. A visualization of the resulting DAN
can be seen in Fig. \ref{fig:Summary-of-proposed} and a single control
module in Fig. \ref{fig:Controller-modules}. 
\begin{center}
\begin{table*}
\begin{centering}
{\small{}}%
\begin{tabular}[t]{ccccccccc||c|c}
{\small{}Net} & {\small{}C-10 } & {\small{}GTSR } & {\small{}SVHN } & {\small{}Caltech } & {\small{}Dped } & {\small{}Oglt } & {\small{}Plnk } & \multicolumn{1}{c}{{\small{}Sketch }} & \multicolumn{1}{c}{{\small{}Perf.}} & {\small{}\#par}\tabularnewline
\hline 
\textbf{\small{}VGG-B(S)} & \textcolor{green}{\small{}92.5}{\small{} } & \textcolor{green}{\small{}98.2}{\small{} } & \textcolor{green}{\small{}96.2}{\small{} } & {\small{}88.2 } & {\small{}92.9 } & \textcolor{green}{\small{}86.9}{\small{} } & \textcolor{red}{\small{}74.5}{\small{} } & \textcolor{red}{\small{}69.2}{\small{} } & \textcolor{blue}{\small{}87.32}{\small{} } & \textcolor{black}{\small{}8}\tabularnewline
\hline 
\textbf{\small{}VGG-B(P)} & \textcolor{red}{\small{}93.2}{\small{} } & \textcolor{red}{\small{}99.0}{\small{} } & \textcolor{blue}{\small{}95.8}{\small{} } & \textcolor{red}{\small{}92.6}{\small{} } & \textcolor{green}{\small{}98.7}{\small{} } & {\small{}83.8 } & \textcolor{blue}{\small{}73.2}{\small{} } & {\small{}65.4 } & \textcolor{green}{\small{}87.71}{\small{} } & \textcolor{black}{\small{}8}\tabularnewline
\hline 
\textbf{\small{}$\mathbf{DAN}_{\mathbf{caltech-256}}$} & {\small{}77.9 } & {\small{}93.6 } & {\small{}91.8 } & {\small{}88.2 } & {\small{}93.8 } & {\small{}81.0 } & {\small{}63.6 } & {\small{}49.4 } & {\small{}79.91 } & \textcolor{green}{\small{}2.54}\tabularnewline
\hline 
\textbf{\small{}$DAN_{sketch}$} & {\small{}77.9 } & {\small{}93.3 } & {\small{}93.2 } & {\small{}86.9 } & {\small{}94.0 } & \textcolor{blue}{\small{}85.4}{\small{} } & {\small{}69.6 } & \textcolor{green}{\small{}69.2}{\small{} } & {\small{}83.7 } & \textcolor{green}{\small{}2.54}\tabularnewline
\hline 
\textbf{\small{}$DAN_{noise}$} & {\small{}68.1 } & {\small{}90.9 } & {\small{}90.4 } & {\small{}84.6 } & {\small{}91.3 } & {\small{}80.6 } & {\small{}61.7 } & {\small{}42.7 } & {\small{}76.29 } & \textcolor{red}{\small{}1.76}\tabularnewline
\hline 
\textbf{\small{}$DAN_{imagenet}$} & \textcolor{blue}{\small{}91.6}{\small{} } & \textcolor{blue}{\small{}97.6}{\small{} } & {\small{}94.6 } & \textcolor{green}{\small{}92.2}{\small{} } & \textcolor{red}{\small{}98.7}{\small{} } & {\small{}81.3 } & {\small{}72.5 } & {\small{}63.2 } & {\small{}86.46 } & \textcolor{blue}{\small{}2.76}\tabularnewline
\hline 
\textbf{\small{}$DAN_{imagenet+sketch}$} & {\small{}91.6 } & {\small{}97.6 } & {\small{}94.6 } & \textcolor{blue}{\small{}92.2}{\small{} } & \textcolor{blue}{\small{}98.7}{\small{} } & {\small{}85.4 } & {\small{}72.5 } & \textcolor{blue}{\small{}69.2}{\small{} } & \textbf{\textcolor{red}{\small{}87.76}}{\small{} } & \textcolor{black}{\small{}3.32}\tabularnewline
\hline 
\end{tabular}
\par\end{centering}{\small \par}
\caption{\label{tab:Baseline-performance:-top}Perf: top-1 accuracy (\%, higher
is better) on various datasets and parameter cost (\#par., lower is
better) for a few baselines and several variants of our method. Rows
1,2: independent baseline performance.\emph{ VGG-B}: VGG \cite{simonyan2014very}
architecture B. (S) - trained from scratch. (P) - pre-trained on ImageNet.
Rows 3-7: (ours) controller network performance\emph{; $DAN_{sketch}$}
as a base network outperforms \textbf{$DAN_{caltech-256}$} on most
datasets. A controller network based on random weights ($DAN_{noise}$)
works quite well given that its number of learned parameters is a
fifth of the other methods. $DAN_{imagenet}$: controller networks
initialized from VGG-B model pretrained on ImageNet. $DAN_{imagenet+sketch}$:
selective control network based on both VGG-B(P) \& Sketch. We color
code the \textcolor{red}{first}, \textcolor{green}{second} and \textcolor{blue}{third}
highest values in each column (lowest for \#par). \#par: amortized
number of weights learned to achieve said performance for all tasks
divided by number of tasks addressed (lower is better).}
\end{table*}
\par\end{center}

We denote a network learned for a dataset/task $S$ as $N_{S}$. A
controller learned using $N_{S}$ as a base network will be denoted
as $DAN_{S}$, where DAN stands for \textbf{D}eep \textbf{A}daptation
\textbf{N}etwork and $DAN_{S\rightarrow T}$ means using $DAN_{S}$
for a specific task $T$. While in this work we apply the method to
classification tasks it is applicable to other tasks as well. 

\subsection{Additional Design Choices}

In the following, we mention some additional design choices of possible
ways to augment a network with controllers, as well as provide some
analysis on the incurred cost of parameters. 

\subsubsection{Weaker parametrization}

A weaker variant of our method is one that forces the matrices $W_{l}$
to be diagonal, e.g, only scaling the output of each filter of the
original network. We call this variant ``\emph{diagonal}'' (referring
only to scaling coefficients, such as by a diagonal matrix) and the
full variant of our method ``\emph{linear}'' (referring to a linear
combination of filters). The diagonal variant can be seen as a form
of explicit regularization which limits the expressive power of the
learned representation. While requiring significantly fewer parameters,
it results in poorer classification accuracy, but as will be shown
later, also outperforms regular feature-extraction for transfer learning,
especially in network compression regimes. 

\subsubsection{Multiple Controllers\label{subsec:Multiple-Controllers} }

The above description mentions one base network and one controller
network. However, any number of controller networks can be attached
to a single base network, regardless of already attached ones. In
this case $\alpha$ is extended to one-hot vector of values determined
by another sub-network, allowing each controller network to be switched
on or off as needed.

\subsubsection{Parameter Cost\label{par:Weight-Efficiency}}

The number of new parameters added for each task depends on the number
of filters in each layer and the number of parameters in the fully-connected
layers. As the latter are not reused, their parameters are fully duplicated.
Let $M=C_{o}\times D$ be the filter dimensions for some conv. layer
$\phi_{l}$ where $D=C_{i}\times k\times k$. A controller module
for $\phi_{l}$ requires $C_{o}^{2}$ coefficients for $F_{l}^{a}$
and an additional $C_{o}$ for \textbf{$b_{l}^{a}$}. Hence the ratio
of new parameters w.r.t to the old for $\phi_{l}$ is $\frac{C_{o}\times(C_{o}+1)}{C_{o}\times(D+1)}=\frac{C_{o}+1}{D+1}\approx\frac{C_{o}}{D}$.
Example: for $C_{o}=C_{i}=256$ input and output units and a kernel
size $k=5$ this equals $\frac{256+1}{256\cdot5^{2}+1}\approx0.04$.
In the final architecture we use the total number of weights required
to adapt the convolutional layers $\Phi_{l}$ combined with a new
fully-connected layer amounts to about 13\% of the original parameters.
For VGG-B, this is roughly 21\%. For instance, constructing 10 classifiers
using one base network and 9 controller networks requires $(1+0.13*9)\cdot P$=$2.17\cdot P$
parameters where $P$ is the number for the base network alone, compared
to $10\cdot P$ required to train each network independently. The
cost is dependent on network architecture, for example it is higher
when applied on the VGG-B architecture. While our method can be applied
to any network with convolutional layers, if $C_{o}\ge D$, \ie,
the number of output filters is greater than the dimension of each
input filter, it would only increase the number of parameters. 

\subsection{Limitations \label{subsec:Theoretical-Limitations}}

An immediate question that arises from our formulation concerns the
expressive power of the resulting adapted networks. We provide some
analysis here on this issue. In regular fine-tuning or learning schemes,
the weights of each layer may take on arbitrary values, depending
on the optimization scheme and training data. In contrast, the proposed
method constrains each filter to be a linear combination of the original
filters in the corresponding layer. This may, in general, become a
strong limiting factor on the performance of the network. The filters
$F=f_{1},\dots,f_{k}$ of a layer $l$ can form vectors by flattening
their tensor form. $F$ defines a basis of a subspace of filters from
which the proposed method creates new ones. As our method defines
new filters by applying a linear transformation on $F$, the initial
values thereof can have several effects on the resulting networks.
Two important scenarios are when (a) network expressibility and (b)
network efficiency are affected. We provide examples of when these
scenarios can occur and some experiments to demonstrate them. 
\begin{center}
\begin{figure*}
\begin{centering}
\subfloat[]{\includegraphics[height=0.47\textwidth]{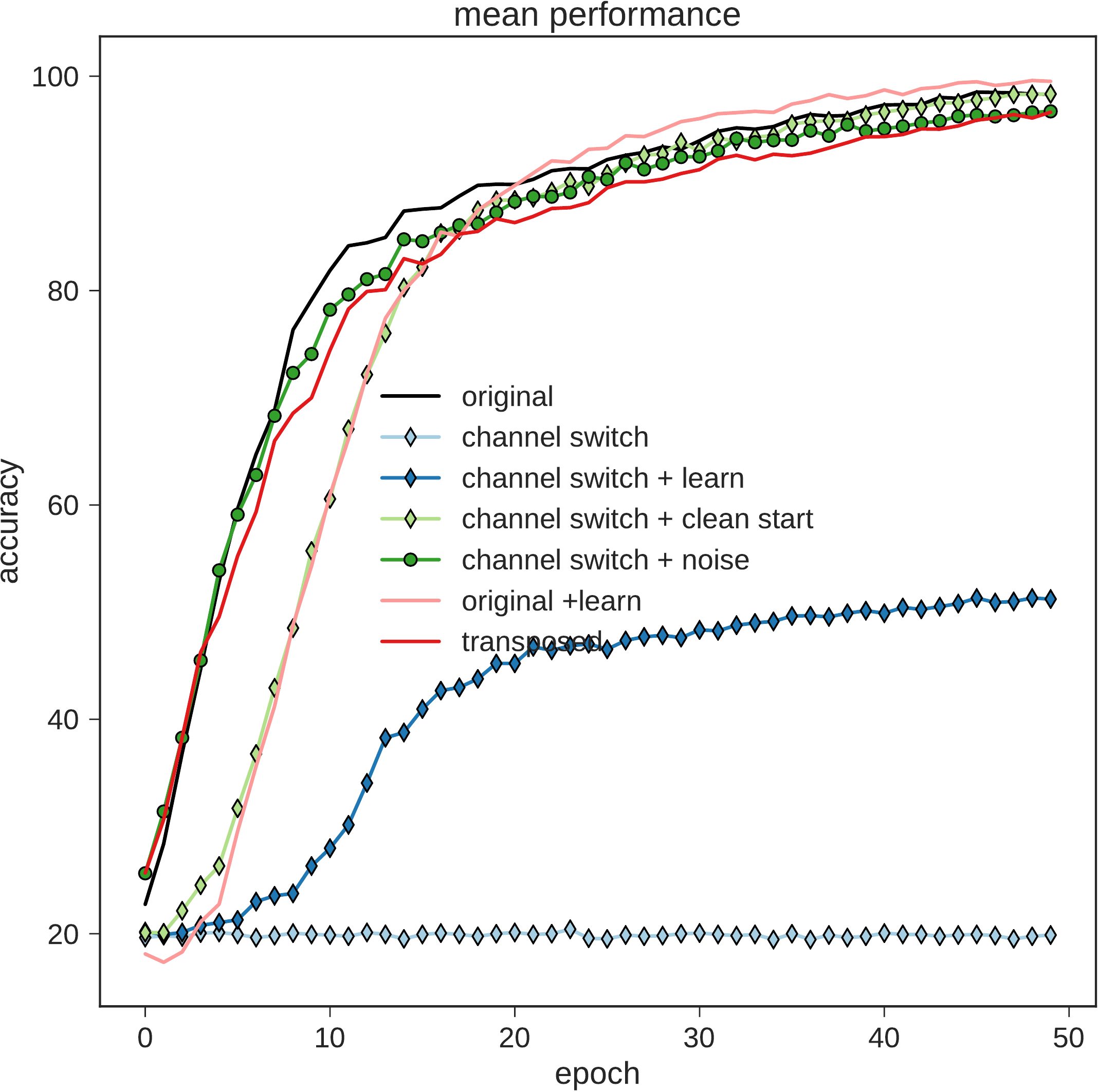}

}\subfloat[]{\includegraphics[height=0.47\textwidth]{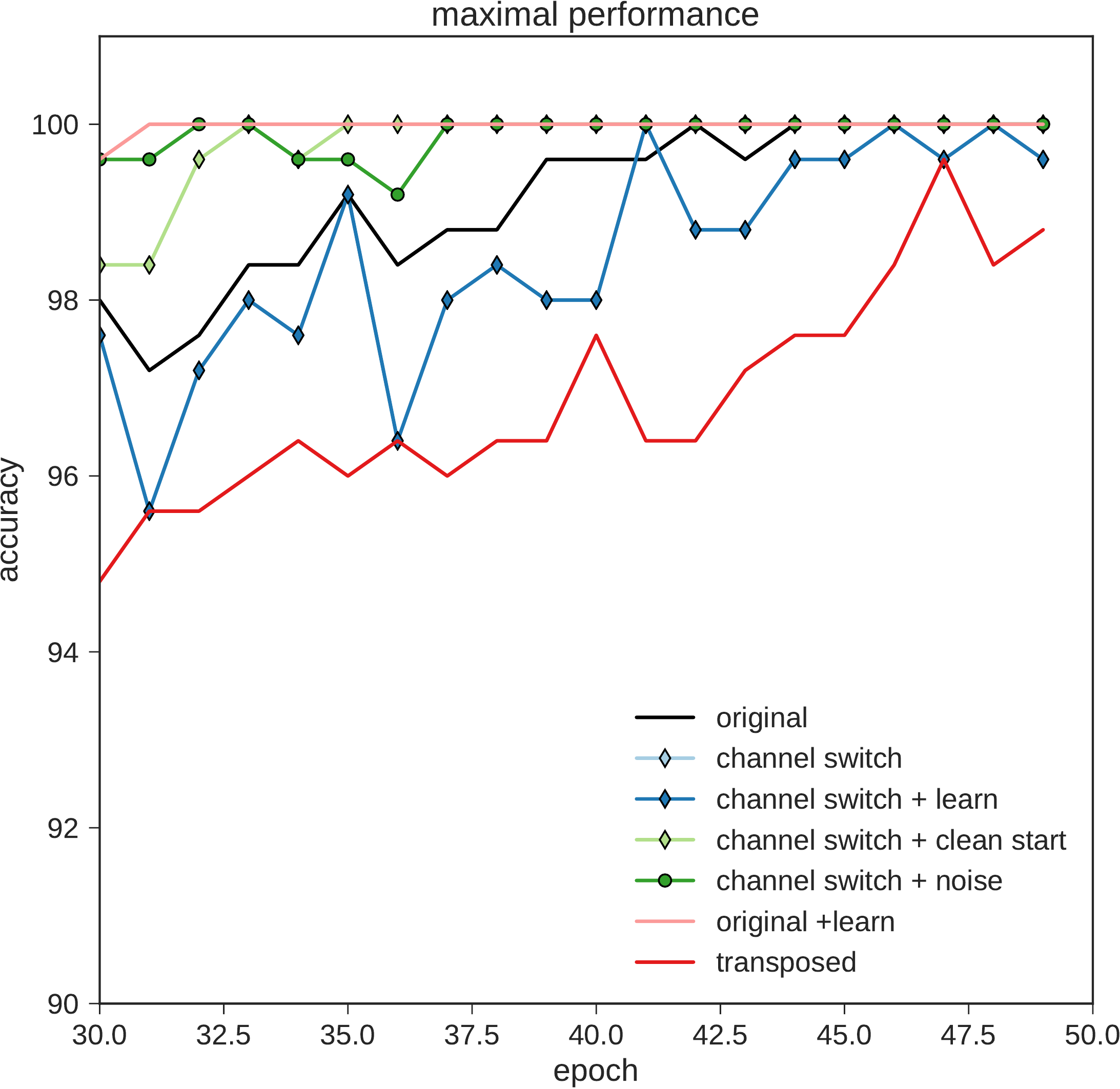}}
\par\end{centering}
\caption{\label{fig:Demonstration-of-different}Demonstration of different
possible limits of possible method on a toy example. Transferring
from a network where the first layer contains features from an orthogonal
sub-space to that required for a task can result in chance accuracy.
Please see text for details. }
\end{figure*}
\par\end{center}

\subsubsection{Expressibility}

One can easily construct cases where the expressive power of a network
is critically hindered by choosing to adapt it from an inappropriate
base network. For example, assume that the inputs of a network are
RGB images. Also assume that - for some reason - all first-layer filters
have coefficients of exactly 0 in the green and blue channels. If
information in these channels is critical for the task to be learned
by the proposed method, it will clearly fail, as the information resides
in an orthogonal feature space to that represented by the base network's
filters. 

\paragraph{Residual Connections}

The above claim does not hold for networks with residual connections
\cite{he2016deep}. In Residual connections the output $y_{i}$ of
the $i'th$ layer is of the following general form:

\begin{equation}
y_{i}=F(x_{i};W_{i})+x_{i}
\end{equation}

where $x_{i}$ is the input to this layer and $W_{i}$ is the layer's
parameters. This allows preserving information from the previous layer,
as a possible function permitted by this is the identity function.
If enough information from the signal $x$ is preserved then the next
layer can still recover from not representing it adequately in this
layer. Hence, the expressive power of the network is probably less
sensitive to the choice of basis function in a single layer. Nevertheless,
its efficiency may be compromised, as we describe next. 

\subsubsection{Efficiency}

In some cases, the expressive power of the network is not immediately
affected by the choice of basis function but we claim that the \emph{efficiency
}of the network is. We define the efficiency of a network w.r.t to
a given image patch as the minimal number of layers (depth) required
to discriminate it from others by a single filter. In other words,
the depth of the shallowest layer where there exists some filter whose
response correlates strongly with the presence of the patch. An example
is a 3x3 gray-scale patch with 1 in all corners and 0 otherwise. In
the first layer, a filter $v$ of values: 
\[
v=\left(\begin{array}{ccc}
+1 & -1 & +1\\
-1 & -1 & -1\\
+1 & -1 & +1
\end{array}\right)
\]
 would perfectly match this patch. On the other hand, one could set
9 filters of the first layer to delta functions with 1 in location
$i,j$ for each $i$ and $j$ in ${\ensuremath{{1,2,3}}}.$ These
filters span all possible patches of 3x3 so they do not effect expressibility.
However, keeping the first layer fixed, the second layer would be
required to create a filter to match $p$, whereas $v$ expressed
this directly. 

\subsubsection{Toy Dataset }

To further demonstrate the above claims, we constructed a toy classification
task. We first describe the dataset and experiments and then discuss
the results.

The task is: given an input image of size 28x28x3, predict the length
of a bar in the an image. In this dataset, the images are restricted
to be 1-pixel width bars of length $3\cdot k$ where $k\in\{3,4,5,6,7\}$
over a blank background and the class is the corresponding length
of the bar. The bar lengths were set to be in increments of 3 as lesser
ones caused difficulties for the network to learn, probably due to
max-pooling operations. 

The dataset is composed of 1000 examples, with a random 75\%/25\%
train/test split. The network we tested is deliberately quite degenerate:
it contains one 5x5 filter on the first layer and 20 filters on the
second, followed by two fully connected layers of 320x50 and 50x5. 

We created 3 variants of this dataset, to test different transfer-learning
scenarios on it. The dataset variants are (1) only red horizontal
bars, (2) only red vertical bars (referred to as ``transposed'')
and (3) only green horizontal bars (referred to as ``new channel'').
Note that we limited our method to operate only on the first layer,
which in this case boils down to multiplying the output of the filter
by a scalar and learning a new bias. 

The network was trained for 50 epochs using the Adam \cite{kingma2014adam}
optimizer (SGD did not converge as well in this case) in all scenarios.
We tested seven scenarios. These involve different combinations of
the images in the dataset, initialization of the first layer and which
layers may or may not be modified by the optimizer. 
\begin{itemize}
\item \textbf{original}: horizontal dataset, first filter fixed to a horizontal
green bar, learn other layers
\item \textbf{original+learn }: horizontal dataset, learn all layers
\item \textbf{transposed} : transposed dataset, first filter fixed to a
horizontal bar, learn other layers
\item \textbf{channel switch}: new channel dataset, first filter fixed to
horizontal green bar, learn other layers
\item \textbf{channel switch + noise}: new channel dataset, first filter
fixed to horizontal green bar+ Gaussian noise, learn other layers
\item \textbf{channel switch + clean start}: new channel dataset, learn
all layers
\item \textbf{channel switch + learn}: new channel dataset, first filter
initialized to a horizontal green bar, learn all layers.
\end{itemize}
Each experiment was repeated 20 times to account for variability.
We plot the mean accuracy on the validation set over the 50 training
epochs. 

Fig. \ref{fig:Demonstration-of-different} summarizes the result of
these experiments. The left sub-figure shows the mean performance
over the 50 epochs and the right one shows the maximal (of the 20
trials) attained for each scenario. These show that in most cases,
we can reach good accuracy, however on the average case there are
drastic differences between the obtained performance, although this
is a very simple dataset. The \textbf{original} scenario, where the
first layer is fixed, attains 100\% accuracy. Switching the channel
from green to red (``\textbf{channel switch}'') but keeping the
first filter fixed results in chance accuracy, as excepted: no linear
combination of a purely ``green'' filter can represent information
from the red channel. In this case, if we allow all layers to be learned
(``\textbf{channel switch + learn}''), indeed the network can finally
converge to a good accuracy (Fig. \ref{fig:Demonstration-of-different}
(b)), but the average performance over 20 trials is far from optimal
(Fig. \ref{fig:Demonstration-of-different} (a)) - the network converges
to around 50\% accuracy (where chance performance is 20\%). This confirms
that even for very simple cases, bad initialization can be detrimental
to the training process. 

As a ``sanity test'', we make sure that randomly initializing the
network and allowing all layers to be learned in the ``channel switch''
case achieves 100\% performance (in both the average and best case).
This is indeed the case (``\textbf{channel switch + clean start}''),
although we see that the random initialization converges at first
much slower than the informed initialization of ``original''. Next,
we initialize the network as in ``\textbf{original}'' but with the
``channel switch'' dataset. However, we add random noise to the
first layer's filter, then keep it fixed (``\textbf{channel switch
+ noise}''). The optimization is able to recover from this initialization
in both the average and best case, as the initial filter is not orthogonal
to the green channel. 

The ``\textbf{original + learn}'' is initialized as ``\textbf{original}''
but allows the first layer to be modified. While on average it initially
converges slower than the ``\textbf{original}'' scenario, on average
it performs better. 

Finally, the ``\textbf{transposed}'' case, where the original filter
is fixed to be orthogonal to the required shape (horizontal bars),
attains average lower performance, and near 100\% in the best case.
Note that although we cannot generate an horizontal bar by linearly
transforming the vertical one, the information is not lost and can
be recovered by filters of the second layer, due to their convolutional
nature. This is done by a weighted summation of the shifted responses
of the horizontal filter with proper coefficients. This is an example
of the network ``efficiency'' being reduced, i.e, delegating to
the second layer computations which could be done in the first. 

To summarize, we conclude from the experiments on this toy dataset
that:
\begin{enumerate}
\item Setting some network weights using strong domain-specific knowledge
and fixing their values, learning only other layers can lead to fast
convergence to a strong solution.
\item An even stronger solution can emerge if all layers are initialized
randomly, then allowed to be updated - though convergence will be
slower. 
\item Initializing some weights to ``bad'' values can cause on-average
dramatically inferior performance to the previous two cases, with
occasional but rare instances of good performance.
\item Initializing some weights to ``bad'' values, fixing them and learning
the rest of the weights will totally and irreparably prevent the network
from learning.
\end{enumerate}
In the next section, we test our method on richer and more diverse
datasets, less degenerate networks (e.g., more than one filter in
the first layer), and in some cases residual networks, which all are
likely to diminish effects shown on artificial examples. 

\section{Experiments on Real Datasets\label{sec:Experiments}}
\begin{center}
\begin{figure*}
\begin{centering}
\subfloat[]{\begin{centering}
\includegraphics[height=0.33\textwidth]{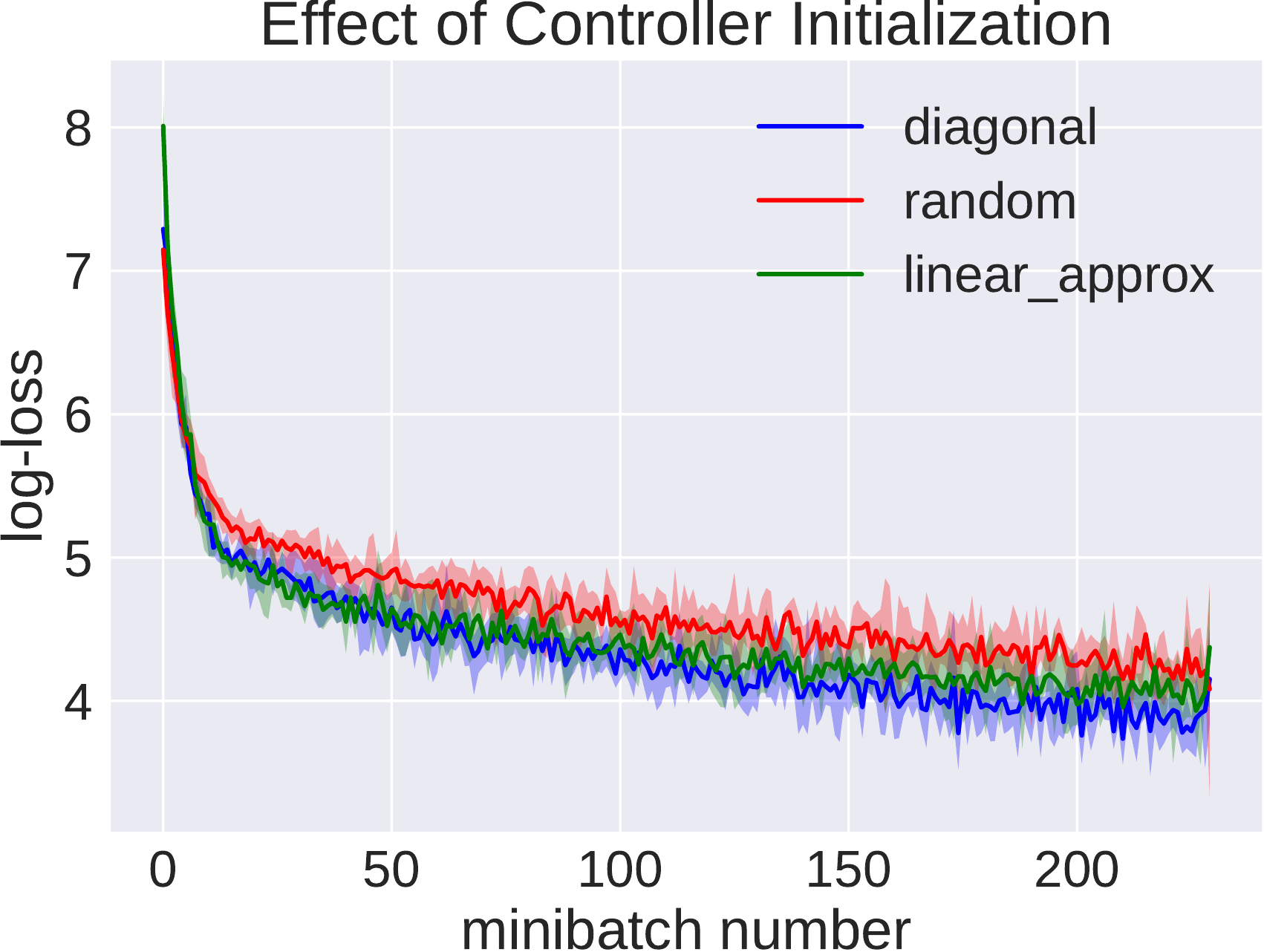}
\par\end{centering}
}\subfloat[]{\begin{centering}
\includegraphics[height=0.33\textwidth]{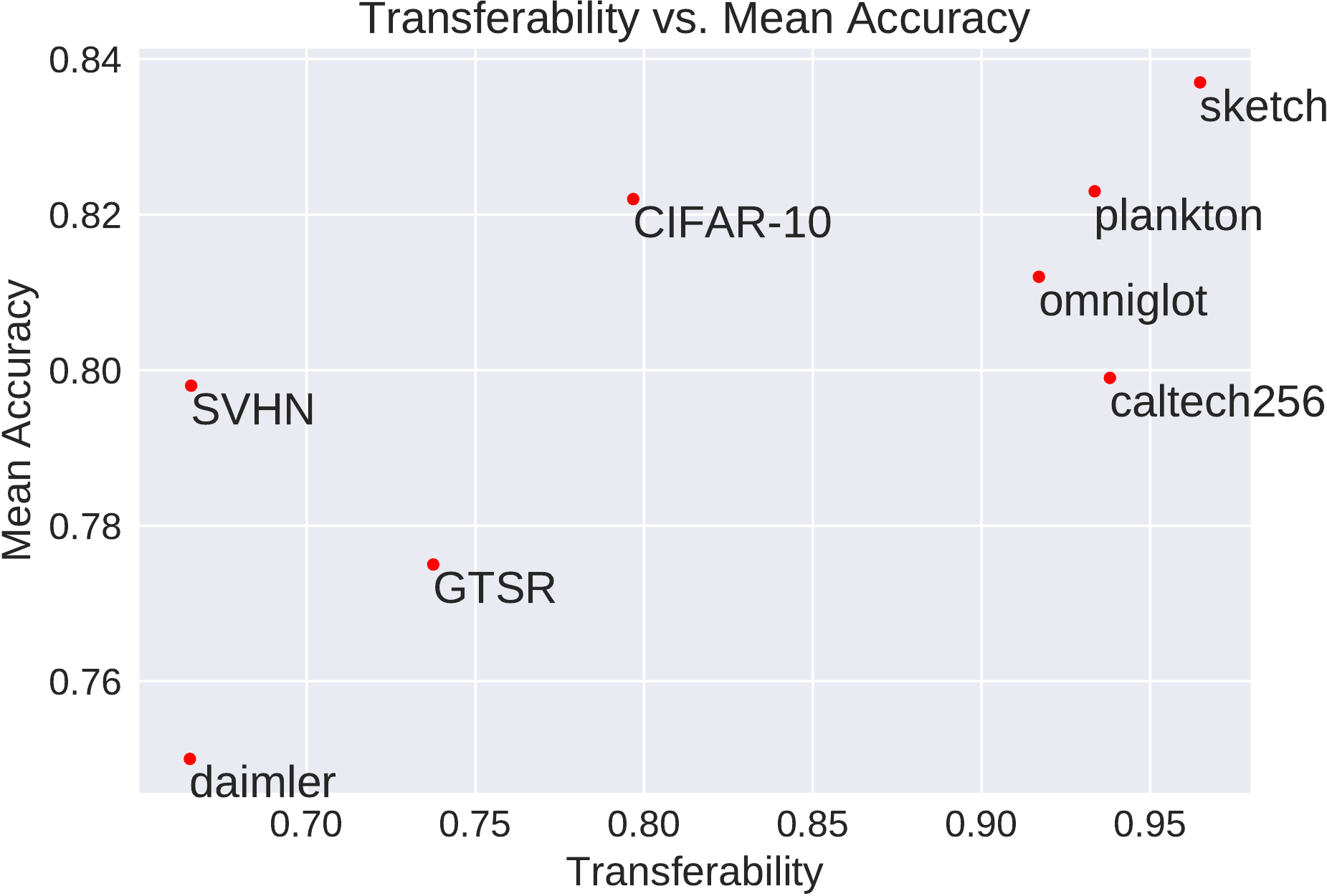}
\par\end{centering}
}
\par\end{centering}
\caption{\label{fig:Controller-initialization-scheme} (\emph{a}) Controller
initialization schemes. Mean loss averaged over 5 experiments for
different ways of initializing controller modules, overlaid with minimal
and maximal values. Random initialization performs the worst (\textcolor{red}{random}).
Approximating the behavior of a fine-tuned network is slightly better
(\textcolor{green}{linear\_approx}) and initializing by mimicking
the base network (\textcolor{blue}{diagonal}) performs the best (\emph{b})
Predictability of a control network's overall accuracy average over
all datasets, given its transferability measure. }
\end{figure*}
\par\end{center}

We conduct several experiments to test our method and explore different
aspects of its behavior. The experimental section is split into two
parts, the first being of a more exploratory nature and the second
geared toward results in a recent multi-task learning benchmark. We
use two different basic network architectures on two (somewhat overlapping)
sets of classification benchmarks for the respective parts. The first
is VGG-B \cite{simonyan2014very} . We begin by listing the datasets
we used (\ref{subsec:Datasets-and-Evaluation}), followed by establishing
baselines by training a separate model for each using a few initial
networks. We proceed to test several variants of our proposed method
(\ref{subsec:Control-Networks}) as well as testing different training
schemes. Next, we discuss methods of predicting how well a network
would fare as a base-network (\ref{subsec:Transferability}). We show
how to discern the domain of an input image and output a proper classification
(\ref{subsec:A-Unified-Network}) without manual choice of the control
parameters $\alpha$. In the second part of our experiments we show
how our applying our method results in leading scores in the Visual
Decathlon Challenge \cite{rebuffi2017learning}, using a different,
more recent architecture. Before concluding we demonstrate some useful
properties of our method. 

All Experiments were performed using the PyTorch\footnote{pytorch.org}
framework using a single Titan-X Pascal GPU. 
\begin{center}
\begin{table*}
\begin{centering}
{\small{}}%
\begin{tabular}[t]{llllllll>{\raggedright}p{0.05\paperwidth}}
{\small{}Dataset } & {\small{}DPed } & {\small{}SVHN } & {\small{}GTSR } & {\small{}C-10 } & {\small{}Oglt } & {\small{}Plnk} & {\small{}Sketch } & {\small{}Caltech }\tabularnewline
\midrule
{\small{}ft-full } & {\small{}60.1 } & {\small{}60 } & {\small{}65.8 } & {\small{}70.9 } & {\small{}80.4 } & {\small{}81.6 } & \textbf{\small{}84.2 } & {\small{}82.3 }\tabularnewline
\midrule
{\small{}ft-full-bn-off } & {\small{}61.8 } & {\small{}64.6 } & {\small{}66.2 } & {\small{}72.5 } & {\small{}78 } & {\small{}80.2 } & \textbf{\small{}82.5 } & {\small{}81 }\tabularnewline
\midrule
{\small{}ft-last } & {\small{}24.4 } & {\small{}33.9 } & {\small{}42.5 } & {\small{}44 } & {\small{}44.1 } & {\small{}47 } & {\small{}50.3 } & \textbf{\small{}55.6 }\tabularnewline
\bottomrule
\end{tabular}
\par\end{centering}{\small \par}
\centering{}\caption{\label{tab:(a)-Mean-transfer}Mean transfer learning performance.
We show the mean top-1 accuracy (\%) attained by fine-tuning a network
from each domain to all domains. Out of the datasets above, starting
with Caltech-256 proves most generic as a feature extractor (\emph{ft-last}).
However, fine tuning is best when initially training on the Sketch
dataset (\emph{ft-full).}}
\end{table*}
\par\end{center}

\subsubsection{Datasets and Evaluation\label{subsec:Datasets-and-Evaluation}}

The first part of our evaluation protocol resembles that of \cite{1701.07275}:
we test our method on the following datasets: \textbf{Caltech-256
}\cite{griffin2007caltech},\textbf{ CIFAR-10} \cite{krizhevsky2009learning},
\textbf{Daimler \cite{munder2006experimental}} (DPed), \textbf{GTSR}
\cite{stallkamp2012man}, \textbf{Omniglot \cite{mnih2015human}},
\textbf{Plankton imagery data \cite{cowen2015planktonset}} ({\small{}Plnk}),
\textbf{Human Sketch dataset} \cite{eitz2012humans} and \textbf{SVHN
\cite{netzer2011reading}}. All images are resized to 64 \texttimes{}
64 pixels, duplicating gray-scale images so that they have 3 channels
as do the RGB ones. We whiten all images by subtracting the mean pixel
value and dividing by the variance per channel. This is done for each
dataset separately. We select 80\% for training and 20\% for validation
in datasets where no fixed split is provided. We use the B architecture
described in \cite{simonyan2014very}, henceforth referred to as VGG-B.
It performs quite well on the various datasets when trained from scratch
(See Tab. \ref{tab:Baseline-performance:-top}. Kindly refer to \cite{1701.07275}
for a brief description of each dataset.

As a baseline, we train networks independently on each of the 8 datasets.
All experiments in this part are done with the Adam optimizer \cite{kingma2014adam},
with an initial learning rate of 1e-3 or 1e-4, dependent on a few
epochs of trial on each dataset. The learning rate is halved after
each 10 epochs. Most networks converge within the first 10-20 epochs,
with mostly negligible improvements afterwards. We chose Adam for
this part due to its fast initial convergence with respect to non-adaptive
optimization methods (e.g, SGD), at the cost of possibly lower final
accuracy \cite{journals/corr/WilsonRSSR17}. The top-1 accuracy (\%)
is summarized in Table \ref{tab:Baseline-performance:-top}.

\subsection{Controller Networks\label{subsec:Control-Networks}}

To test our method, we trained a network on each of the 8 datasets
in turn to be used as a base network for all others. We compare this
to the baseline of training on each dataset from scratch (\textbf{VGG-B(S)})
or pretrained (\textbf{VGG-B(P)}) networks. Tab. \ref{tab:Baseline-performance:-top}
summarizes performance on all datasets for two representative base
nets: $DAN_{caltech-256}$ (79.9\%) and $DAN_{sketch}$ (83.7\%).
Mean performance for other base nets are shown in Fig. \ref{fig:Controller-initialization-scheme}
(a). The parameter cost (\ref{par:Weight-Efficiency}) of each setting
is reported in the last column of the table. This (similarly to \cite{rebuffi2017learning}
is the total number of parameters required for a set of tasks normalized
by that of a single fully-parametrized network. We also check how
well a network can perform as a base-network after it has seen ample
training examples: $DAN_{imagenet}$ is based on VGG-B pretrained
on ImageNet \cite{russakovsky2015imagenet}. This improves the average
performance by a significant amount (83.7\% to 86.5\%). On Caltech-256
we see an improvement from 88.2\% (training from scratch). However,
for both Sketch and Omniglot the performance is in favor of $DAN_{sketch}$.
Note these are the only two domains of strictly unnatural images.
Additionally, $DAN_{imagenet}$ is still slightly inferior to the
non-pretrained \textbf{VGG-B(S)} (86.5\% vs 87.7\%), though the latter
is more parameter costly. 

\subsubsection{Multiple Base Networks}

Arguably, a good base network should have features generic enough
so that a controller network can use them for a broad range of target
tasks. In practice this may not necessarily be the case. We conjecture
that using a diverse set of features, such as that learned by training
on more than one task, will provide a better basis for transfer learning.
To use two base-networks simultaneously, we implemented a dual-controlled
network by using both $DAN_{caltech-256}$ and $DAN_{sketch}$ and
attaching to them controller networks. The outputs of the feature
parts of the resulting sub-networks were concatenated before the fully-connected
layer. This resulted in the exact same performance as $DAN_{sketch}$
alone. However, by using selected controller-modules per group of
tasks, we can improve the results: for each dataset the maximally
performing network (based on validation) is the basis for the control
module, \ie, we used $DAN_{imagenet}$ for all datasets except Omniglot
and Sketch. For the latter two we use $DAN_{sketch}$ as a base net.
We call this network $DAN_{imagenet+sketch}$. At the cost of more
parameters, it boosts the mean performance to \textbf{87.76}\% - better
than using any single base net for controllers or training from scratch.
Since it is utilized for 9 tasks (counting ImageNet), its parameter
cost (2.76) is still quite good.

\subsubsection{Starting from a Randomly Initialized Base Network}

We tested how well our method can perform without any prior knowledge,
\eg, building a controller network on a \emph{randomly initialized
base network. }The total number of parameters for this architecture
is \textasciitilde{}12M. However,\emph{ }as 10M have been randomly
initialized and only the controller modules and fully-connected layers
have been learned, the effective number is actually 2M. Hence its
parameter cost is determined to be 0.22. We summarize the results
in Tab. \ref{tab:Baseline-performance:-top}. Notably, the results
 of this initialization worked surprisingly well; the mean top-1 precision
attained by this network was 76.3\%, slightly worse than of $DAN_{caltech-256}$
(79.9\%). This is better than initializing with $DAN_{daimler}$,
which resulted in a mean accuracy of 75\%. This is possible because
the random values in the base network can still be linearly combined
by our method to create ones that are useful for classification.

\subsection{Initialization }

One question which arises is how to initialize the weights $W$ of
a control-module. We tested several options: (1) Setting $W$ to an
identity matrix (\textbf{diagonal}). This is equivalent to the controller
module starting with a state which effectively mimics the behavior
of the base network (2) Setting $W$ to random noise (\textbf{random)
}(3) Training an independent network for the new task from scratch,
then set $W$ to best linearly approximate the new weights with the
base weights (\textbf{linear\_approx}). To find the best initialization
scheme, we trained $DAN_{sketch\rightarrow caltech256}$ for one epoch
with each and observed the loss. Each experiment was repeated 5 times
and the results averaged. From Fig. \ref{fig:Controller-initialization-scheme}(a),
it is evident that the \textbf{diagonal} initialization is superior,
perhaps counter-intuitively, there is no need to train a fully parametrized
target network. Simply starting with the behavior of the base network
and tuning it via the control modules results in faster convergence.
Hence we train controller modules with the diagonal method. Interestingly,
the residual adaptation unit in \cite{rebuffi2017learning} is initially
similar to the diagonal configuration. If all of the filters in their
adapter unit are set to 1 (up to normalization), the output of the
adapter will be initially the same as that of the controller unit
initialized with the identity matrix.

\subsection{Transferability\label{subsec:Transferability}}
\begin{center}
\begin{figure}
\begin{centering}
\includegraphics[width=1\columnwidth]{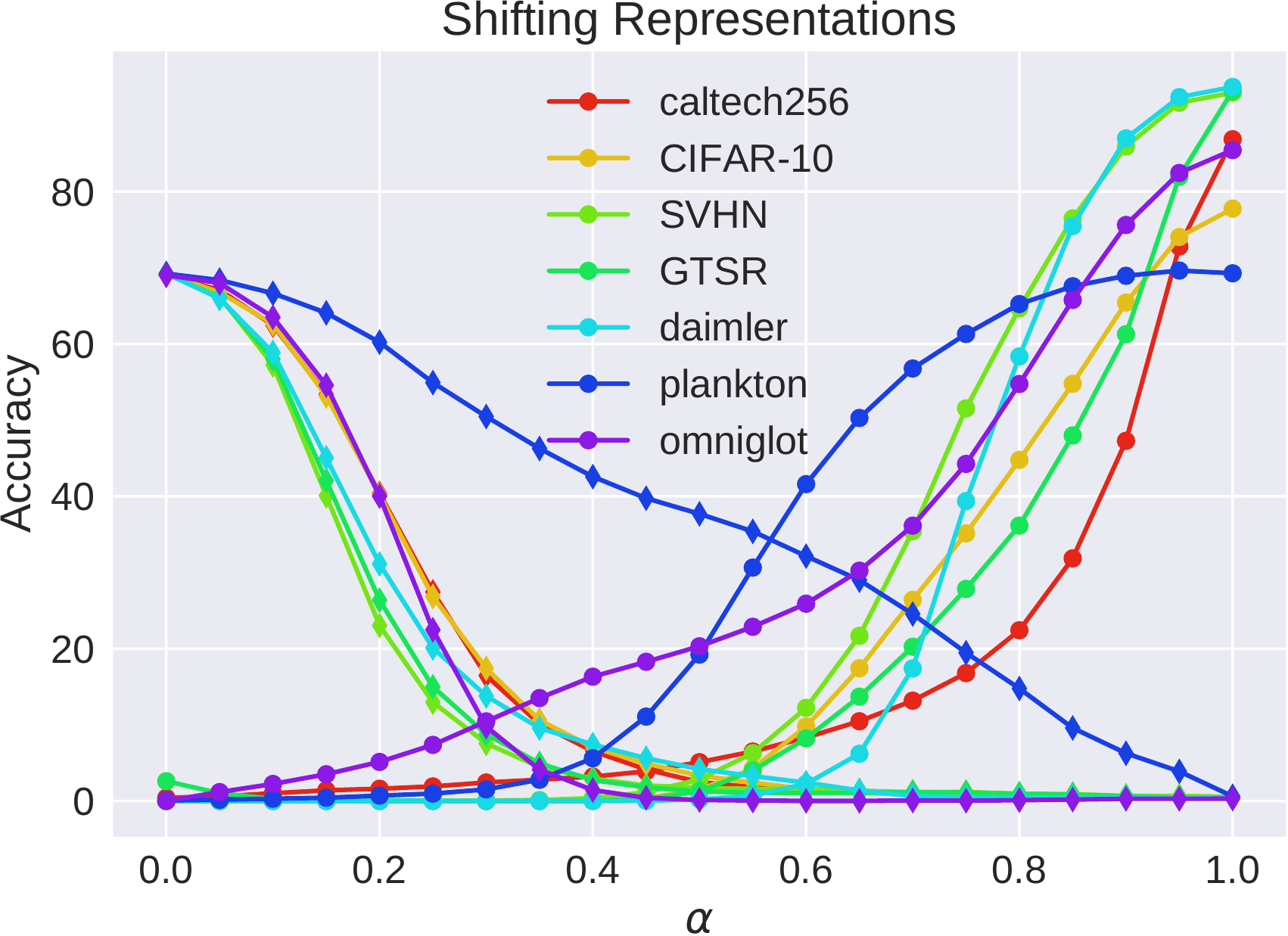}
\par\end{centering}
\caption{\label{fig:Shifting-Representations.-Using}Shifting Representations.
Using a single base network $N_{sketch}$, we check the method's sensitivity
to varying values of $\alpha$ by varying it in the range $[0,1]$.
Increasing $\alpha$ shifts the network away from the base representation
and towards learned tasks - gradually lowering performance on the
base task (diamonds) and improving on the learned ones (full circles).
The relatively slow decrease of the performance on sketch (blue diamonds)
and increase in that of Plankton (blue circles) indicates a similarity
between the learned representations.}
\end{figure}
\par\end{center}

\begin{center}
\begin{figure*}
\begin{centering}
\subfloat[]{\includegraphics[height=0.25\textheight]{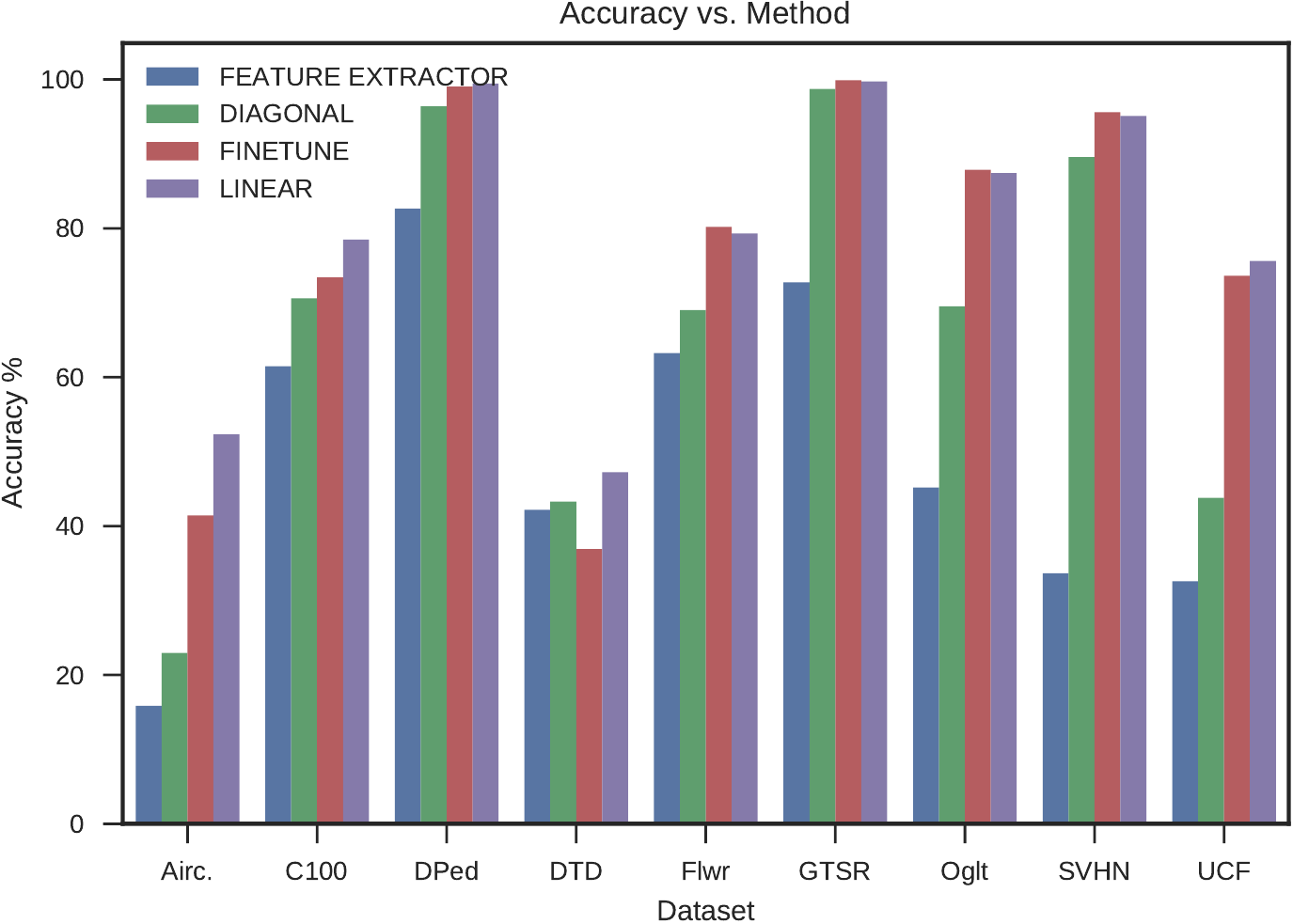}}\subfloat[]{\includegraphics[height=0.25\textheight]{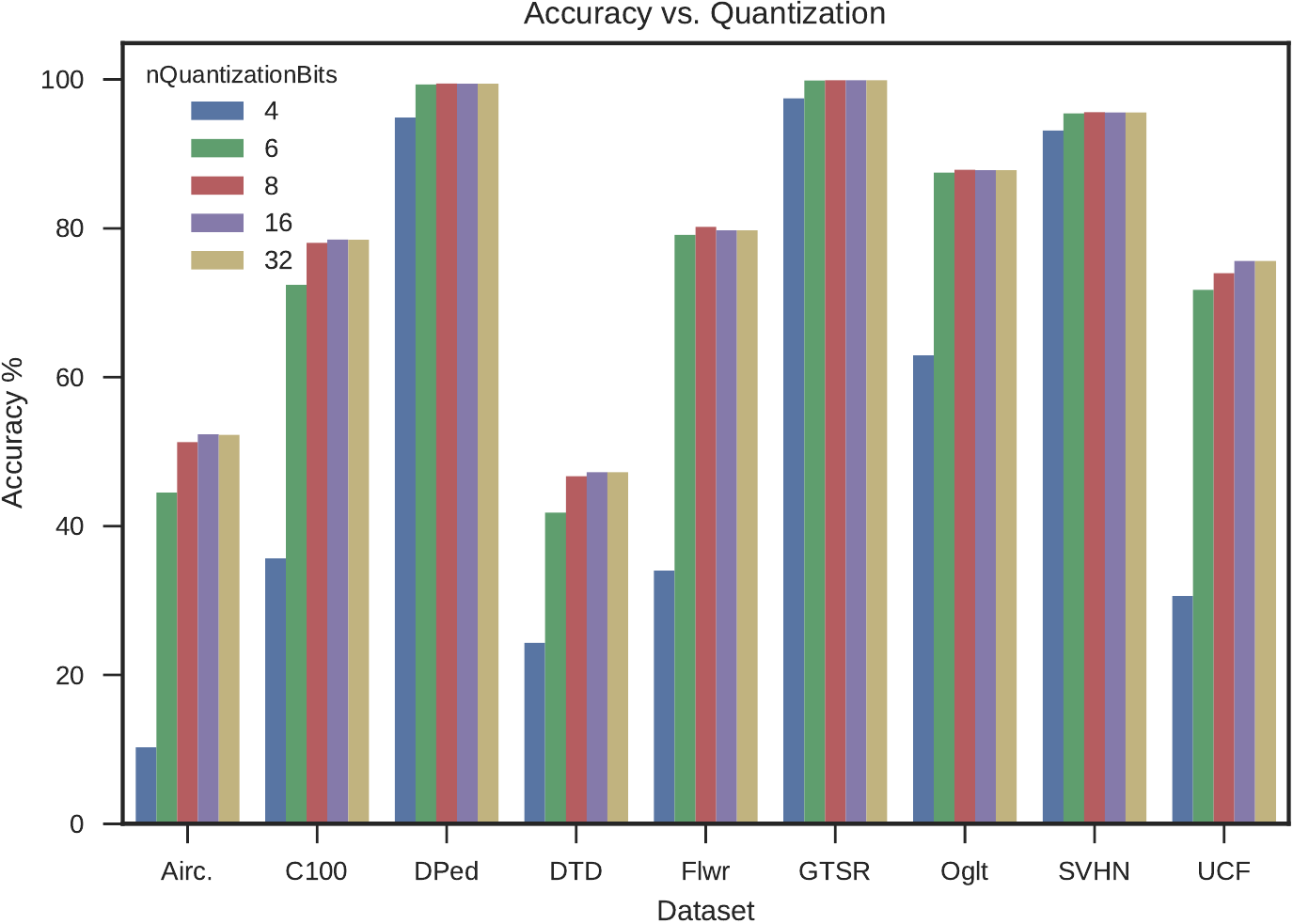}}
\par\end{centering}
\caption{\label{fig:(a)-Accuracy-vs.}(a) Accuracy vs. learning method. Using
only the last layer (\emph{feature extractor}) performs worst. \emph{finetune}:
vanilla fine-tuning. \emph{Diagonal} : our controller modules with
a diagonal combination matrix. \emph{Linear}: our full method. On
average, our full method outperforms vanilla fine tuning. (b) Accuracy
vs. quantization: with as low as 8 bits, we see no significant effect
of network quantization on our method, showing they can be applied
together. }
\end{figure*}
\par\end{center}

How is one to choose a good network to serve as a base-network for
others? As an indicator of the representative power of the features
of each independently trained network $N$, we test the performance
on other datasets, using $N$ for fine tuning. We define the \emph{transferability}
of a source task $S$ w.r.t a target task $T$ as the top-1 accuracy
attained by fine-tuning $N$ trained on $S$ to perform on $T$. We
test 3 different scenarios, as follows: (1)\textbf{ }Fine-tuning only
the last layer (a.k.a feature extraction) \textbf{(ft-last}); (2)
Fine-tuning all layers of $N$(\textbf{ft-full)}; (3) same as ft-full,
but freezing the parameters of the batch-normalization layers - this
has proven beneficial in some cases - we call this option \textbf{ft-full-bn-off}.
The results in Fig \ref{fig:Transferability-of-various} show some
interesting phenomena. First, as expected, feature extraction (ft\_last)
is inferior to fine-tuning the entire network. Second, usually training
from scratch is the most beneficial option. Third, we see a distinction
between natural images (Caltech-256, CIFAR-10, SVHN, GTSR, Daimler)
and unnatural ones (Sketch, Omniglot, Plankton); Plankton images are
essentially natural but seem to exhibit different behavior than the
rest. 

It is evident that features from the natural images are less beneficial
for the unnatural images. Interestingly, the converse is not true:
training a network starting from Sketch\emph{ }or Omniglot\emph{ }works
quite well for most datasets, both natural and unnatural. This is
further shown in Tab. \ref{tab:(a)-Mean-transfer}: we calculate the
mean transferability of each dataset by the mean value of each rows
of the transferability matrix from Fig. \ref{fig:Transferability-of-various}.
$DAN_{Caltech-256}$ works best for feature extraction. However, for
full fine-tuning using $DAN_{Plankton}$ works as the best starting
point, closely followed by $DAN_{Caltech-256}$. For controller networks,
the best mean accuracy attained for a single base net trained from
scratch is attained using $DAN_{sketch}$ (83.7\%). This is close
to the performance attained by full transfer learning from the same
network (84.2\%, see Tab. \ref{tab:(a)-Mean-transfer}) at a fraction
of the number of parameters. This is consistent with our transferability
measure. To further test the correlation between the transferability
and the performance given a specific base network, we used each dataset
as a base for control networks for all others and measured the mean
overall accuracy. The results can be seen in Fig. \ref{fig:Controller-initialization-scheme}
(b). 

\subsubsection{A Unified Network\label{subsec:A-Unified-Network}}

Finally, we test the possibility of a single network which can both
determine the domain of an image and classify it. We train a classifier
to predict from which dataset an image originates, using the training
images from the 8 datasets. This is learned easily by the network
(also VGG-B) which rapidly converges to 99\% accuracy. With this ``dataset-decider'',
named $N_{dc}$ we augment $DAN_{sketch}$ to set for each input image
$I$ from any of the datasets $D_{i}$ the controller value $\alpha_{i}$
of $DAN_{sketch\rightarrow D_{i}}$ to 1 if and only if $N_{dc}$
deemed $I$ to originate from $D_{i}$ and to 0 otherwise. This produces
a network which applies to each input image the correct controllers,
classifying it within its own domain. While the predicted values of
$\alpha_{i}$ are real-valued, we set the highest one to 1 and the
rest to zero. 

\subsection{Visual Decathlon Challenge \label{subsec:Visual-Decathlon-Challenge}}

We now show results on the recent Visual Decathlon Challenge of \cite{rebuffi2017learning}.
The challenge introduces involves 10 different image classification
datasets: \textbf{ImageNet} \cite{russakovsky2015imagenet}; \textbf{Aircraft}
\cite{maji2013fine}; \textbf{Cifar-100} \cite{krizhevsky2009learning};
\textbf{Daimler Pedestrians} \cite{munder2006experimental}; \textbf{Dynamic
Textures} \cite{journals/corr/CimpoiMKMV13}; \textbf{GTSR} \cite{stallkamp2012man};
\textbf{Flowers} \cite{conf/icvgip/NilsbackZ08}; \textbf{Omniglot}
\cite{lake2015human}; \textbf{SVHN} \cite{netzer2011reading} and\textbf{
UCF-101} \cite{1212.0402v1}. The goal is to achieve accurate classification
on each dataset while retaining a small model size, using the train/val/test
splits fixed by the authors. All images are resized so the smaller
side of each image is 72 pixels. The classifier is expected to operate
on images of 64x64 pixels. Each entry in the challenge is assigned
a decathlon score, which is a function designed to highlight methods
which do better than the baseline on all 10 datasets. Please refer
to the challenge website for details about the scoring and datasets:
\url{http://www.robots.ox.ac.uk/~vgg/decathlon/}. Similarly to \cite{rebuffi2017learning},
we chose to use a wide residual network \cite{journals/corr/ZagoruykoK16}
with an overall depth of 28 and a widening factor of 4, with a stride
of 2 in the convolution at the beginning of each basic block. In what
follows we describe the challenge results, followed by some additional
experiments showing the added value of our method in various settings.
In this section we used the recent YellowFin optimizer \cite{1706.03471v1}
as it required less tuning than SGD. We use an initial learning rate
factor of 0.1 and reduce it to 0.01 after 25 epochs. This is for all
datasets with the exception of ImageNet which we train for 150 epochs
with SGD with an initial learning rate of 0.1 which is reduced every
35 epochs by a factor of 10. This is the configuration we determined
using the available validation data which was then used to train on
the validation set as well (as did the authors of the challenge) and
obtain results from the evaluation server. Here we trained on the
reduced resolution ImageNet from scratch and used the resulting net
as a base for all other tasks. Tab. \ref{tab:visual-decathlon-results}
summarizes our results as well as baseline methods and the those of
\cite{rebuffi2017learning}, all using a base architecture of similar
capacity. By using a significantly stronger base architecture they
obtained higher results (mean of 79.43\%) but with a parameter cost
of 12, i.e., requiring 12 times the amount of original parameters.
All of the rows are copied from \cite{rebuffi2017learning}, including
their re-implementation of LWF, except the last which shows our results.
The final column of the table shows the decathlon score. A score of
2500 reflects the baseline resulting from finetuning the network learned
on ImageNet to each dataset independently. For the same architecture,
the best results obtained by the Residual Adapters method is slightly
 below ours in terms of decathlon score and slightly above them in
terms of mean performance. However, unlike them, we avoid joint training
over all of the datasets and using dataset-dependent weight decay. 
\begin{center}
\begin{table*}
\begin{centering}
{\small{}}%
\begin{tabular}{>{\raggedright}m{0.15\columnwidth}ccccccccccc||cc}
{\small{}Method} & {\small{}\#par} & {\small{}ImNet} & {\small{}Airc.} & {\small{}C100} & {\small{}DPed} & {\small{}DTD} & {\small{}GTSR} & {\small{}Flwr} & {\small{}Oglt} & {\small{}SVHN} & {\small{}UCF} & {\small{}mean} & {\small{}S}\tabularnewline
\hline 
{\small{}Scratch} & {\small{}10} & {\small{}59.87 } & {\small{}57.1 } & {\small{}75.73 } & {\small{}91.2 } & {\small{}37.77 } & {\small{}96.55 } & {\small{}56.3 } & {\small{}88.74 } & {\small{}96.63 } & {\small{}43.27 } & {\small{}70.32 } & {\small{}1625 }\tabularnewline
\hline 
\hline 
{\small{}Feature} & {\small{}1} & {\small{}59.67 } & {\small{}23.31 } & {\small{}63.11 } & {\small{}80.33 } & {\small{}45.37 } & {\small{}68.16 } & {\small{}73.69 } & {\small{}58.79 } & {\small{}43.54 } & {\small{}26.8 } & {\small{}54.28 } & {\small{}544 }\tabularnewline
\hline 
\hline 
{\small{}Finetune} & {\small{}10} & {\small{}59.87 } & {\small{}60.34 } & {\small{}82.12 } & {\small{}92.82 } & {\small{}55.53 } & {\small{}97.53 } & {\small{}81.41 } & {\small{}87.69 } & {\small{}96.55 } & {\small{}51.2 } & {\small{}76.51 } & {\small{}2500 }\tabularnewline
\hline 
\hline 
{\small{}LWF} & {\small{}10} & {\small{}59.87 } & {\small{}61.15 } & {\small{}82.23 } & {\small{}92.34 } & {\small{}58.83 } & {\small{}97.57 } & {\small{}83.05 } & {\small{}88.08 } & {\small{}96.1 } & {\small{}50.04 } & {\small{}76.93 } & {\small{}2515 }\tabularnewline
\hline 
\hline 
{\small{}Res. Adapt. } & {\small{}2} & {\small{}59.67 } & {\small{}56.68 } & {\small{}81.2 } & {\small{}93.88 } & {\small{}50.85 } & {\small{}97.05 } & {\small{}66.24 } & {\small{}89.62 } & {\small{}96.13 } & {\small{}47.45 } & {\small{}73.88 } & {\small{}2118 }\tabularnewline
\hline 
\hline 
{\small{}Res. Adapt (Joint)} & {\small{}2} & {\small{}59.23 } & {\small{}63.73 } & {\small{}81.31 } & {\small{}93.3 } & {\small{}57.02 } & {\small{}97.47 } & {\small{}83.43 } & {\small{}89.82 } & {\small{}96.17 } & {\small{}50.28 } & \textbf{\small{}77.17 } & {\small{}2643 }\tabularnewline
\hline 
\hline 
{\small{}DAN (Ours)} & {\small{}2.17} & {\small{}57.74} & {\small{}64.12} & {\small{}80.07} & {\small{}91.3} & {\small{}56.54 } & {\small{}98.46} & {\small{}86.05} & {\small{}89.67 } & {\small{}96.77} & {\small{}49.38} & {\small{}77.01} & \textbf{\small{}2851}\tabularnewline
\hline 
\end{tabular}
\par\end{centering}{\small \par}
\caption{\label{tab:visual-decathlon-results}Results on Visual Decathlon Challenge.
\emph{Scratch}: training on each task independently. \emph{Feature}:
using a pre-trained network as a feature extractor. \emph{Finetune}
: vanilla fine tuning. Performs well but requires many parameters.
Learning-without-forgetting (\emph{LWF}, \cite{journals/corr/LiH16e}
slightly outperforms it but with a large parameter cost. \emph{Residual
adapt.} \cite{rebuffi2017learning} significantly reduce the number
of parameters. Results improve when training jointly on all task (\emph{Res.Adapt(Joint)}).
The proposed method (DAN) outperforms residual adapters despite adding
each task \emph{independently} of the others. \textbf{S} is the decathlon
challenge score. }
\end{table*}
\par\end{center}

\subsection{Compression and Convergence}
\begin{center}
\begin{figure*}
\begin{centering}
\subfloat[]{\includegraphics[width=1\columnwidth]{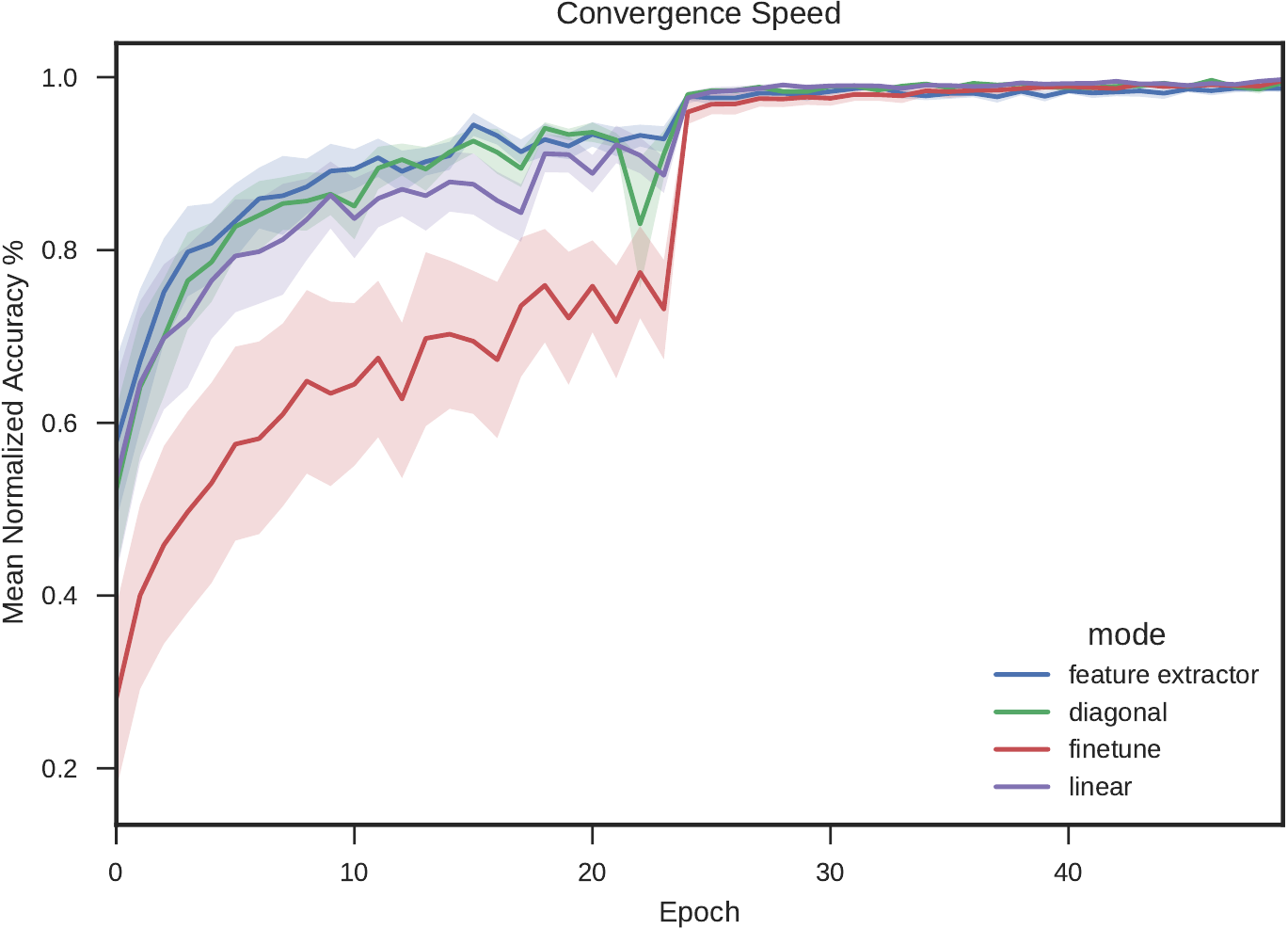}

}\subfloat[]{\includegraphics[width=1\columnwidth]{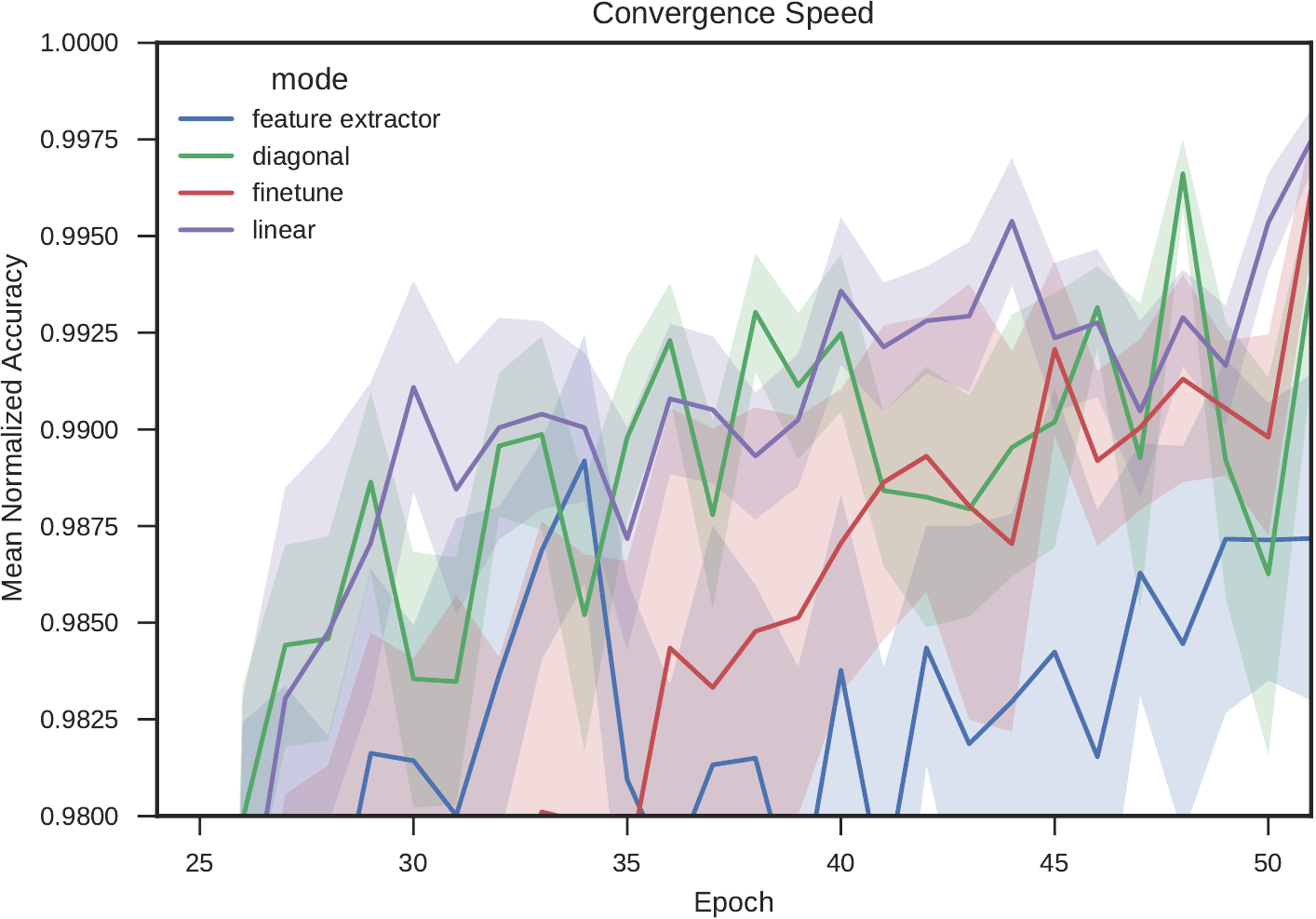}

}
\par\end{centering}
\caption{\label{fig:(a)-Our-method}(a) Our method (\emph{linear}) converges
to a high accuracy faster than fine-tuning. The weaker variant of
our method converges as fast as feature-extraction but reaches an
overall higher accuracy (\ref{fig:(a)-Accuracy-vs.} (a)). (b) zoom
in on top-right of (a). }
\end{figure*}
\par\end{center}

\begin{figure}
\includegraphics[width=1\columnwidth]{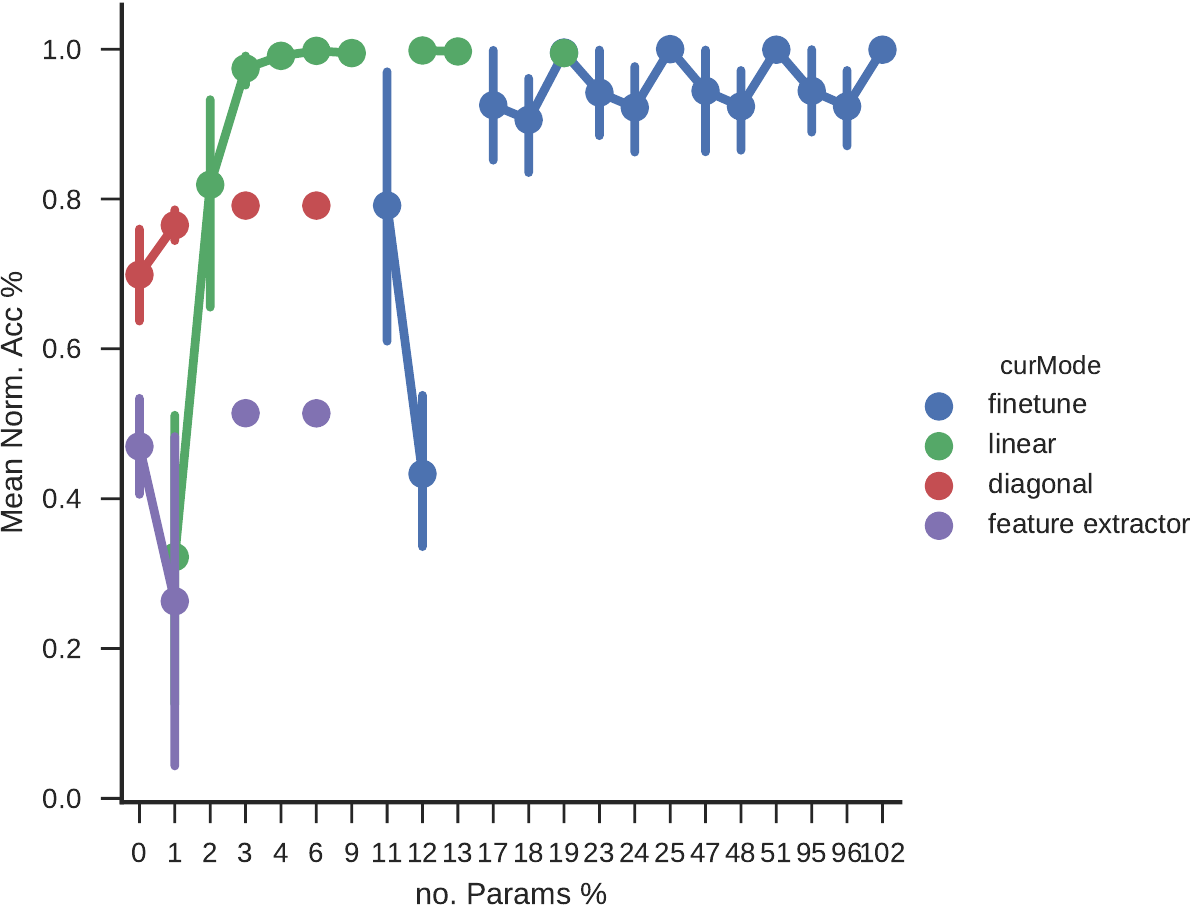}

\caption{\label{fig:Mean-classification-quantization}Mean classification accuracy
(normalized, averaged over datasets) w.r.t no. parameters. Our method
achieve better performance over baselines for a large range of parameter
budgets. For very few parameters diagonal (ours) outperforms features
extraction. To obtain maximal accuracy our full method requires far
fewer parameters (see linear vs finetune). }

\end{figure}

In this section, we highlight some additional useful properties of
our method. All experiments in the following were done using the same
architecture as in the last section but training was performed only
on the training sets of the Visual Decathlon Challenge and tested
on the validation sets. First, we check whether the effects of network
compression are complementary to ours or can hinder them. Despite
the recent trend of sophisticated network compression techniques (for
example \cite{han2015deep}) we use only a simple method of compression
as a proof-of-concept, noting that using recent compression methods
will likely produce better results. We apply a simple linear quantization
on the network weights, using either 4, 6, 8, 16 or 32 bits to represent
each weight, where 32 means no quantization. We do not quantize batch-normalization
coefficients. Fig. \ref{fig:(a)-Accuracy-vs.} (b) shows how accuracy
is affected by quantizing the coefficients of each network. Using
8 bits results in only a marginal loss of accuracy. This effectively
means our method can be used to learn new tasks while requiring an
addition of 3.25\% of the original amount of parameters. Many maintain
performance even at 6 bits (DPed, Flowers, GTSR, Omniglot, SVHN. Next,
we compare the effect of quantization on different transfer methods:
feature extraction, fine-tuning and our method (both diagonal and
linear variants). For each dataset we record the normalized (divided
by the max.) accuracy for each method/quantization level (which is
transformed into the percentage of required parameters). This is plotted
in Fig. \ref{fig:Mean-classification-quantization}. Our method requires
significantly fewer parameters to reach the same accuracy as fine-tuning.
If parameter usage is limited, the diagonal variant of our method
significantly outperforms feature extraction. Finally, we show that
the number of epochs until nearing the maximal performance is markedly
lower for our method. This can be seen in Fig. \ref{fig:(a)-Our-method}. 

\subsection{Discussion}

We have observed that the proposed method converges to a reasonably
good solution faster than vanilla fine-tuning and eventually attains
slightly better performance. The increase in classification accuracy
is despite the network's expressive power, which is limited by our
construction. We conjecture that constraining each layer to be expressed
as a linear combination of the corresponding layer in the original
network serves to regularize the space of solutions and is beneficial
when the tasks are sufficiently related to each other. One could come
up with simple examples where the proposed method would likely fail.
This might occur if tasks are not of the same kind. For example, one
task requires counting of horizontal lines and the other requires
counting of vertical ones, and such examples are all that appear in
the training sets, then the proposed method will likely work far worse
than vanilla fine-tuning or training from scratch. We experimented
with such scenarios in Sec. \ref{subsec:Theoretical-Limitations}.
Empirically, we did not see this happen in any of our experiments
on ``real'' datasets. This may simply stem from some inherent similarity
in image classification tasks which makes some basic set of features
useful for most of them.  We leave the investigation of this issue,
as well as finding ways between striking a balance between reusing
features and learning new ones as future work. 

\section{Conclusions}

We have presented a method for transfer learning thats adapts an existing
network to new tasks while fully preserving the existing representation.
Our method matches or outperforms vanilla fine-tuning, though requiring
a fraction of the parameters, which when combined with net compression
reaches 3\% of the original parameters with no loss of accuracy. The
method converges quickly to high accuracy while being on par or outperforming
other methods with the same goal. Built into our method is the ability
to easily switch the representation between the various learned tasks,
enabling a single network to perform seamlessly on various domains.
The control parameter $\alpha$ can be cast as a real-valued vector,
allowing a smooth transition between representations of different
tasks. An example of the effect of such a smooth transition can be
seen in Fig. \ref{fig:Shifting-Representations.-Using} where $\alpha$
is used to linearly interpolate between the representation of differently
learned pairs tasks, allowing one to smoothly control transitions
between different behaviors. Allowing each added task to use a convex
combination of already existing controllers will potentially utilize
controllers more efficiently and decouple the number of controllers
from the number of tasks. 
\ifCLASSOPTIONcompsoc
  \section*{Acknowledgments}
\else
  \section*{Acknowledgment}
\fi

This research was funded by the Canada Research Chairs program and by the Air Force Office for Scientific Research (USA) for which the authors are grateful.

\bibliographystyle{IEEEtran}
\bibliography{bare_jrnl_compsoc}

\ifCLASSOPTIONcaptionsoff
  \newpage
\fi

 \begin{wrapfigure}{l}{25mm} 
    \includegraphics[width=1in,height=1.25in,clip,keepaspectratio]{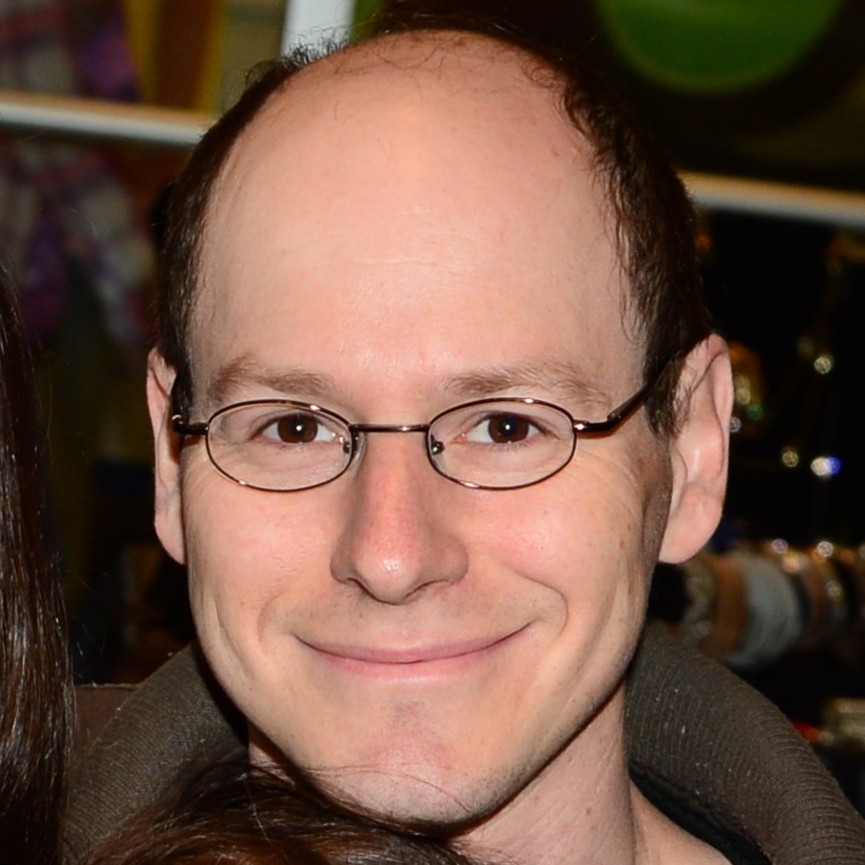}
  \end{wrapfigure}\par
  \textbf{Amir Rosenfeld} received his B.Sc in Computer-Engineering at the Hebrew University of Jerusalem (HUJI) at 2006
and his M.Sc. in Computer Science at HUJI in 2011. He completed his PhD in Computer Science at the Weizmann Institute under the supervision of Prof. Shimon Ullman in 2016. He is currently a postdoctoral fellow at the Laboratory for Active and Attentive Vision, York University. His main research interests include machine learning, computer and human vision, visual attention and adaptive learning systems.\par
\vspace{10px}

\begin{wrapfigure}{l}{25mm} 
    \includegraphics[width=1in,height=1.25in,clip,keepaspectratio]{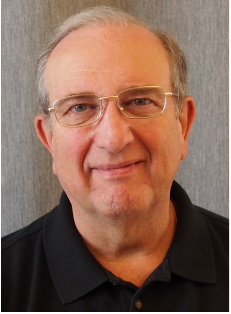}
  \end{wrapfigure}\par
  \textbf{John K. Tsotsos} is Distinguished Research Professor of Vision Science at York University. He received
his doctorate in Computer Science from the University
of Toronto. After a postdoctoral fellowship in
Cardiology at Toronto General Hospital, he joined
the University of Toronto on faculty in Computer
Science and in Medicine. In 1980 he founded the
Computer Vision Group at the University of Toronto,
which he led for 20 years. He was recruited to
York University in 2000 as Director of the Centre
for Vision Research. He has been a Canadian Heart
Foundation Research Scholar, Fellow of the Canadian Institute for Advanced
Research and Canada Research Chair in Computational Vision. He received
many awards and honours including several best paper awards, the 2006
Canadian Image Processing and Pattern Recognition Society Award for
Research Excellence and Service, the 1st President’s Research Excellence
Award by York University in 2009, and the 2011 Geoffrey J. Burton Memorial
Lectureship from the United Kingdom’s Applied Vision Association for
significant contribution to vision science. He was elected as Fellow of the Royal Society of Canada in 2010 and was awarded its 2015 Sir John William Dawson Medal for sustained excellence in multidisciplinary research, the first computer scientist to be so honoured. Over 125 trainees have passed through his lab. His current research focuses on a comprehensive theory of visual attention in humans. A practical outlet for this theory embodies elements of the theory into the vision systems of mobile robots.\par
\end{document}